\documentclass{article} 
\usepackage{iclr2026_conference,times}

\usepackage{amsmath,amsfonts,bm}

\def\eqref#1{equation~\ref{#1}}

\def\1{\bm{1}}

\DeclareMathAlphabet{\mathsfit}{\encodingdefault}{\sfdefault}{m}{sl}
\SetMathAlphabet{\mathsfit}{bold}{\encodingdefault}{\sfdefault}{bx}{n}

\usepackage{hyperref}
\usepackage{url}
\usepackage{booktabs}
\usepackage{graphicx}
\usepackage{xspace}
\usepackage{amsmath}
\usepackage{multirow}
\usepackage{diagbox}
\usepackage{array}
\usepackage[table]{xcolor}
\usepackage[most]{tcolorbox}  
\usepackage[table]{xcolor} 
\usepackage{colortbl}    
\usepackage{tabularx}
\usepackage{enumitem}
\usepackage{tipa}
\usepackage{cleveref}

\title{Teaching LLMs to Abstain via Fine-Grained Semantic Confidence Reward}

\author{%
}
\author{
Hao An \ \
Yang Xu  \\ \\
Computational Linguistics and Consciousness Sciences Lab, \\ 
Southern University of Science and Technology, China \\
\texttt{anh2024@mail.sustech.edu.cn, xuyang@sustech.edu.cn}
}

\newcommand{\Ours}{\textsc{FiSCoRe}\xspace}

\begin{document}

\maketitle

\begin{abstract}
Mitigating hallucinations in Large Language Models (LLMs) is critical for their reliable deployment. Existing methods typically fine-tune LLMs to abstain from answering questions beyond their knowledge scope. However, these methods often rely on coarse-grained signals to guide LLMs to abstain, such as overall confidence or uncertainty scores on multiple sampled answers, which may result in an imprecise awareness of the model's own knowledge boundaries. To this end, we propose a novel reinforcement learning framework built on $\textbf{\underline{Fi}ne-grained \underline{S}emantic \underline{Co}nfidence \underline{Re}ward (\Ours)}$, which guides LLMs to abstain via sample-specific confidence. Specifically, our method operates by sampling multiple candidate answers and conducting semantic clustering, then training the LLM to retain answers within high-confidence clusters and discard those within low-confidence ones, thereby promoting accurate post-hoc abstention. Additionally, we propose a new metric for evaluating the reliability of abstention fine-tuning tasks more comprehensively. Our method significantly enhances reliability in both in-domain and out-of-distribution benchmarks. 
\end{abstract}

\section{Introduction}
Large language models (LLMs) have achieved remarkable success, demonstrating powerful capabilities in content generation, complex reasoning, and software development \citep{gpt4, llama3, qwen2.5}. Their widespread adoption has established them as essential tools in numerous applications. However, a critical vulnerability threatens their reliability: their propensity to hallucinate, fabricating plausible but false information when faced with questions beyond their knowledge \citep{zhang2023siren}. Such hallucinations erode user trust and can propagate misinformation. Reliance on fabricated content in high-stakes domains such as medicine or law can have severe consequences. Therefore, mitigating hallucinations is essential for the safe and responsible deployment of LLMs.

A promising direction for mitigating hallucination is to fine-tune LLMs to abstain from answering questions that lie beyond their knowledge scope and answer those within it~\citep{abstention_survey,cheng2024can}. Supervised methods often categorize a training set into ``known'' and ``unknown'' questions based on the correctness or accuracy of the model's answers, then train the model to answer the former and abstain from the latter \citep{R-tuning,cheng2024can}. To avoid reliance on ground-truth labels, other approaches leverage uncertainty metrics such as semantic entropy to partition the training set into known and unknown questions \citep{R-tuning, SE_Tuning, ualign}. However, these methods rely on coarse, global signals, such as a single correctness label or an aggregated uncertainty score, to guide abstention decisions. These approaches don't account for the nuanced confidence levels of individual samples, resulting in an ambiguous decision boundary. Consequently, models may either excessively abstain from answerable questions or fail to abstain when facing low-confidence inputs, which prevents the model from balancing truthfulness and helpfulness.

\begin{figure}[h]
\begin{center}
\includegraphics[width=0.8\textwidth]{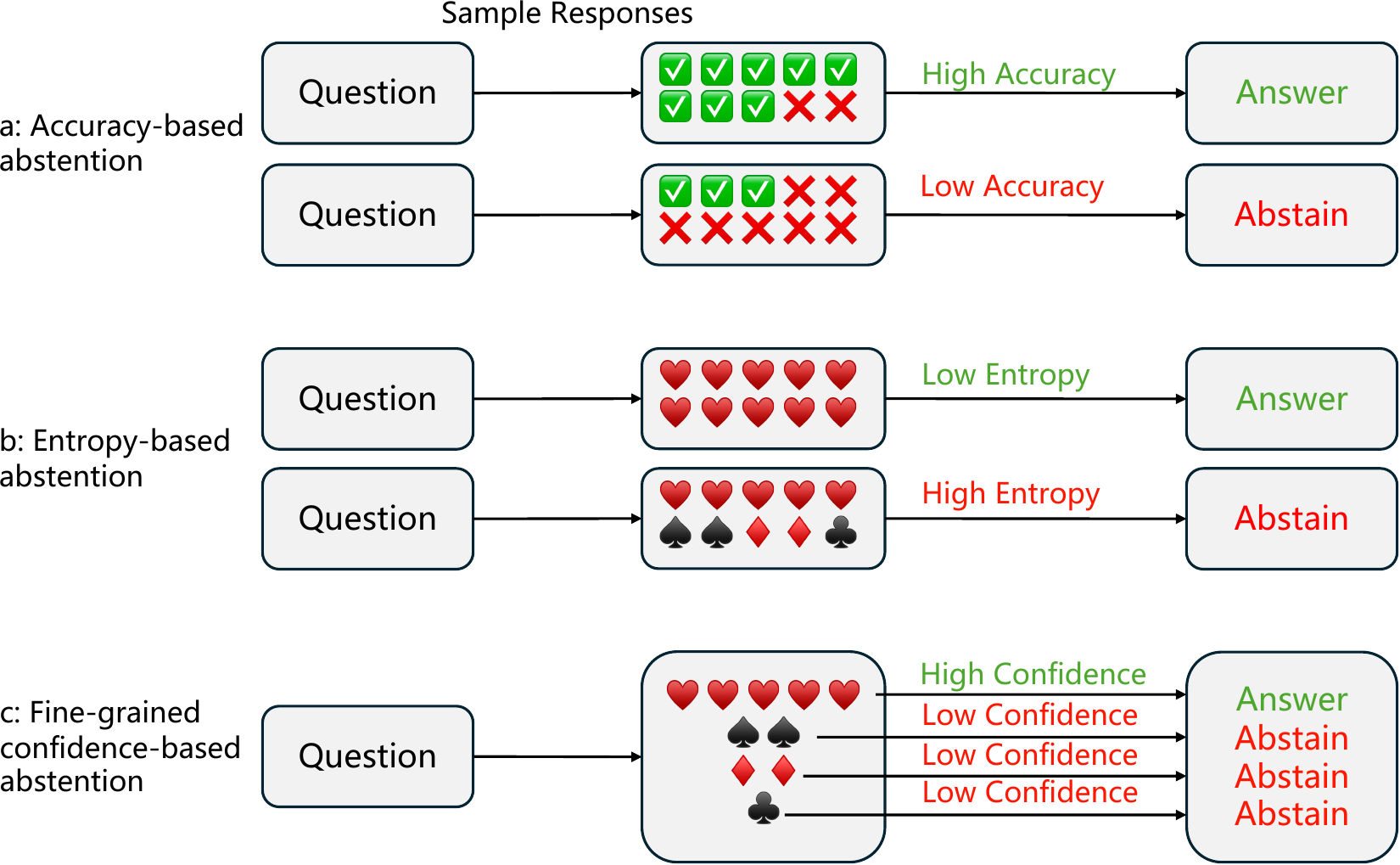}
\end{center}
\caption{(a) Accuracy-based abstention. (b) Entropy-based abstention. (c) Ours: fine-grained confidence-based abstention.}
\label{figure_intro}
\end{figure}

To overcome these limitations, we propose \textbf{Fine-grained Semantic Confidence Reward (\Ours, \textipa{/\'fIs.kO:r/})}, a novel reinforcement learning framework that replaces the conventional global uncertainty reward with a fine-grained, \textbf{per-sample} confidence reward. The core insight is that an answer's intrinsic confidence is reflected in its semantic consensus among all samples, and such confidence can be fully utilized for factual alignment. Although existing methods also sample multiple answers, they typically distill them into a single aggregated uncertainty score. In contrast, as shown in Figure~\ref{figure_intro}, our approach rewards the model for accurately assessing the confidence of each sample. Specifically, we first generate multiple candidate responses and group them into clusters based on their semantic equivalence. We use the cluster size as a proxy for the intrinsic confidence of its member samples: a large cluster implies high confidence, whereas a small cluster implies low confidence. A positive reward is granted only when the model's expressed confidence (e.g., ``sure'' or ``unsure'') aligns correctly with this intrinsic confidence level. This fine-grained signal incentivizes the model to align its verbalized confidence with semantic consensus, thus enabling it to learn a more nuanced knowledge boundary and make more accurate abstention decisions.

Existing studies indicate that helpfulness and truthfulness are like two ends of a seesaw — enhancing one often leads to a decline in the other~\citep{zhu-etal-2025grait}. We notice that current metrics fail to accurately reflect both aspects, making them inadequate for assessing abstention methods. Regarding this, we propose a reliability metric grounded in the model's self-awareness of its knowledge boundaries. This metric synthesizes helpfulness and truthfulness, aiming to prevent models from over-rejecting queries or generating hallucinations.

Our main contributions are as follows:
\vspace{-.5em}
\begin{itemize}[leftmargin=1.2em]
    \item We design a fine-grained semantic confidence reward (see \Cref{sec:reward_func_design}), which significantly improves the reliability of LLM.
    
    \item We propose a new metric (see \Cref{rel_metric_analyze}) that integrates the helpfulness and truthfulness of LLM into a single reliability metric, offering a comprehensive assessment of its awareness of its own knowledge boundaries.

    \item We experiment with various LLMs and test them on both in-distribution and out-of-distribution data. The results demonstrate that our approach outperforms baselines in most cases, particularly on out-of-distribution data, verifying the effectiveness and generalization capability of the proposed fine-grained semantic confidence reward.
\end{itemize}

\section{Problem Formulation}
\subsection{Abstention Fine-Tuning}
To mitigate the risk of hallucination inherent in LLMs, a primary alignment strategy is to fine-tune models to selectively abstain from answering questions that fall beyond their knowledge boundaries~\citep{abstention_survey,li-etal-2025-knowledge-boundary}. The objective is to produce a model that is both helpful, by correctly answering questions within its knowledge boundary, and truthful, by abstaining from those outside it. To formally measure progress towards this goal, we establish an evaluation framework that assesses the performance of a refined LLM relative to the initial LLM before fine-tuning. As illustrated in the abstention confusion matrix in Table~\ref{tab:abstention_confusion_matrix}, we partition the evaluation benchmark into two distinct categories based on the performance of the initial LLM:
\begin{itemize}[leftmargin=1.2em]
    \item \textbf{Known Questions}: The set of questions that the initial LLM answers correctly.
    \item \textbf{Unknown Questions}: The set of questions that the initial LLM answers incorrectly.
\end{itemize}

The refined LLM is then evaluated on its ability to preserve accuracy on known questions while learning to abstain from unknown questions.

\begin{table}[t!]
    \centering
    \caption{Abstention confusion matrix.}
    \begin{center}

    \begin{tabular}{l|c|c|c}
        \toprule
        \diagbox{Initial LLM}{Refined LLM} & Correctly answered & Incorrectly answered & Abstained \\
        \midrule
        Known Questions  & $N_1$ & $N_2$ & $N_3$ \\
        Unknown Questions &  --   & $N_4$ & $N_5$ \\
        \bottomrule
    \end{tabular}
    \end{center}
    \label{tab:abstention_confusion_matrix}
\end{table}

\subsection{Reliability Evaluation}
\label{rel_metric_analyze}

\subsubsection{Helpfulness and Truthfulness}
\label{subsec:current_metrics}

Helpfulness evaluates a model’s capacity to preserve accuracy on questions within its known scope, preventing unnecessary abstention and forgetting. Truthfulness measures its ability to abstain from questions outside that scope, preventing fabricated responses. Following~\citet{kim-etal-2024-aligning,feng-etal-2024-dont}, we adopt two fine-grained F1 metrics $\text{F1}_{ans}$ for helpfulness evaluation and $\text{F1}_{abs}$ for truthfulness evaluation:
\begin{align}
\text{F1}_{ans}
&= \frac{2 \cdot \text{Precision}_{ans}\cdot \text{Recall}_{ans}}
       {\text{Precision}_{ans} + \text{Recall}_{ans}}
= \frac{2 N_1}{2 N_1 + 2 N_2 + N_3 + N_4},\label{eq:F1ans}\\
\text{Precision}_{ans} &= \frac{N_1}{N_1 + N_2 + N_4},\quad
\text{Recall}_{ans} = \frac{N_1}{N_1 + N_2 + N_3},\\
\text{F1}_{abs}
&= \frac{2 \cdot \text{Precision}_{abs}\cdot \text{Recall}_{abs}}
       {\text{Precision}_{abs} + \text{Recall}_{abs}}
= \frac{2 N_5}{N_3 + N_4 + 2 N_5},\label{eq:F1abs}\\
\text{Precision}_{abs} &= \frac{N_5}{N_3 + N_5},\quad
\text{Recall}_{abs} = \frac{N_5}{N_4 + N_5}.
\end{align}

\subsubsection{Reliability Metric}
\label{subsec:proposed_metric}
Since increasing truthfulness often leads to a decrease in helpfulness, using either metric alone introduces bias. Therefore, we propose \(\text{F1}_{rel}\), defined as the harmonic mean of \(\text{F1}_{ans}\) and \(\text{F1}_{abs}\), which provides a holistic measure of reliability:
\begin{equation}
\text{F1}_{rel} = \frac{2 \cdot \text{F1}_{ans} \cdot \text{F1}_{abs}}{\text{F1}_{ans} + \text{F1}_{abs}} = \frac{4 N_1 N_5}{4 N_1 N_5 + 2 N_2 N_5 + N_1 N_3 + N_1 N_4 + N_3 N_5 + N_4 N_5}.
\end{equation}
The \(\text{F1}_{rel}\) metric holistically considers all categories in the abstention confusion matrix in Table~\ref{tab:abstention_confusion_matrix} and exhibits the following desiderata: it monotonically increases with the correctly answered known questions ($N_1$) and correct abstention from unknown questions ($N_5$); it decreases monotonically with both types of error: incorrect answer ($N_2$, $N_3$) and over-abstention ($N_4$), and therefore is intuitively reasonable.  
In the ideal case (\(N_2 = N_3 = N_4 = 0\), \(N_1 + N_5 = N\)), \(\text{F1}_{rel} = 1\); in the worst case (\(N_1 = N_5 = 0\), \(N_2 + N_3 + N_4 = N\)), \(\text{F1}_{rel} = 0\), where $N$ is the number of all questions. This metric effectively balances helpfulness and truthfulness, mitigating the risks of overly conservative or aggressive model behavior.
A recent metric, reliability score (RS) \citep{xurejection}, has been proposed to achieve similar balancing effect, but we show that it is flawed in encouraging hallucination (see \Cref{app:reliability_score}).

\section{Method}
Inspired by the powerful generalization capabilities of reinforcement learning~\citep{sft_vs_rl} and its success in reasoning~\citep{guo2025deepseek-r1}, we incorporate abstention decisions into the rollout process and train LLMs to abstain appropriately via fine-grained confidence reward (\Ours). In this section, we first show the details of training \Ours and formalize the reinforcement learning objective. Then, we demonstrate the prompt template that guides the LLMs to generate rollout in the defined format. Finally, we introduce the reward modeling to guide the optimization of reinforcement learning to improve reliability.

\begin{figure}[h]
\begin{center}
\includegraphics[width=0.9\textwidth]{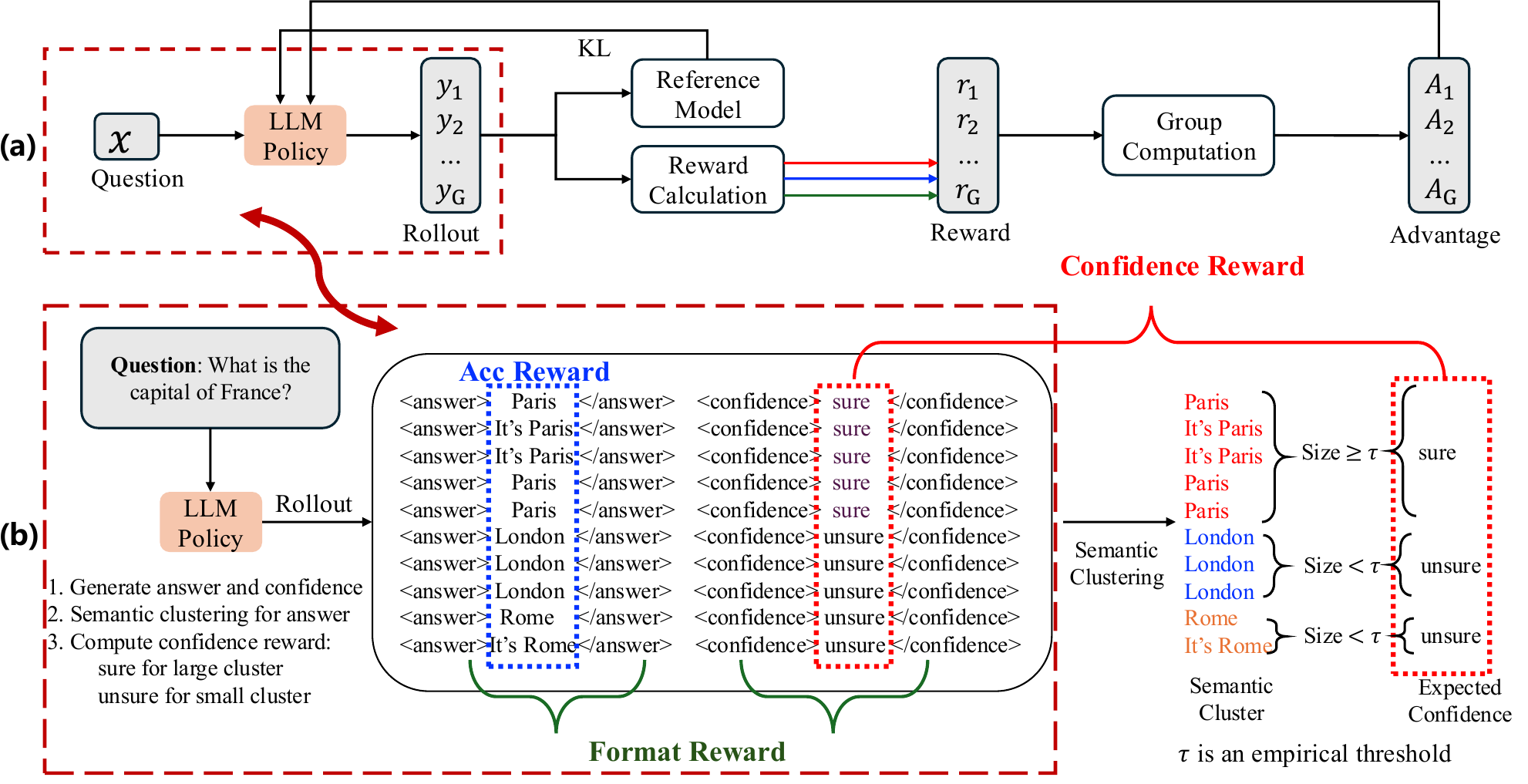}

\end{center}
\caption{The training overview of our method. (a) The GRPO pipeline. (b) A detailed example of how \Ours works.}
\label{method}
\end{figure}

\subsection{Group Relative Policy Optimization}
We reward the model based on the sample-specific semantic confidence among a group of its own sampled responses, which requires a reinforcement learning algorithm capable of handling multiple samples per prompt. Group Relative Policy Optimization (GRPO)~\citep{shao2024deepseekmath} naturally fits this setting: it samples a group of $G$ outputs (or \textit{rollouts}) for each question $x$, which directly supports our semantic clustering–based confidence reward computation. Formally, for each question \( x \), GRPO samples a group of \( G \) outputs \( Y=\{y_1, y_2, \dots, y_G\} \) from the old policy \( \pi_{\theta_{\text{old}}} \) and optimizes the new policy \( \pi_{\theta} \) by maximizing the following objective:
\begin{equation}
\begin{split}
    \mathcal{J}_{\text{GRPO}}(\theta) &= \mathbb{E}_{[x \sim P(Q), \{y_i\}_{i=1}^G \sim \pi_{\theta_{\text{old}}}(Y|x)]} \\
    & \frac{1}{G} \sum_{i=1}^G \left( \min \left( \frac{\pi_\theta(y_i | x)}{\pi_{\theta_{\text{old}}}(y_i | x)} A_i, \text{clip} \left( \frac{\pi_\theta(y_i | x)}{\pi_{\theta_{\text{old}}}(y_i | x)}, 1 - \epsilon, 1 + \epsilon \right) A_i \right) - \beta \mathbb{D}_{\text{KL}}(\pi_{\theta} || \pi_{\text{ref}}) \right),
\end{split}
\label{eq:grpo_obj}
\end{equation}
where $A_{i} = \left(r_i - \text{mean}(\{r_i\}_{i=1}^{G})\right) / \text{std}(\{r_i\}_{i=1}^{G})$ is the normalized advantage of the $i$th rollout in the current group, $\epsilon$ is the clipping ratio, and $\beta$ is the KL loss coefficient. Moreover, the objective also incorporates a KL divergence penalty, specifically to prevent excessive divergence of the policy from the original reference policy model. The GRPO pipeline is shown in Figure~\ref{method} (a).

\subsection{Training Template}
\begin{table*}[h]
\caption{Training template. The question is appended at the end during training and inference.}
\begin{center}
    
\begin{tabular}{p{13cm}}
\toprule
You are a helpful and truthful AI Assistant. Given a question, you should provide responses that include answer and confidence. You first answer the question as briefly as possible enclosed by \verb|<answer>| and \verb|</answer>|, and then provide your confidence in sure or unsure about the answer, enclosed by \verb|<confidence>| and \verb|</confidence>|.  Respond in the following format:

  \textcolor{blue}{\texttt{<answer>}}
  ...
  \textcolor{blue}{\texttt{</answer>}}

  \textcolor{red}{\texttt{<confidence>}}
  sure or unsure
  \textcolor{red}{\texttt{</confidence>}}
  
  Question:\\
\bottomrule
\end{tabular}
\end{center}

\label{template}
\end{table*}
As we describe the confidence reward in Figure~\ref{method}~(b), it is essential for the policy LLM to generate responses that conform to the specified format. To facilitate this, we design a prompt template to guide the policy model in producing an answer along with an associated confidence level. As shown in Table~\ref{template}, the model is instructed to format its responses into two distinct parts. First, the model provides the answer wrapped in the \verb|<answer> </answer>| tag. Second, the model expresses a binary confidence level (sure or unsure) wrapped in the \verb|<confidence> </confidence>| tag. This \verb|<answer> + <confidence>| structure enforces the model to reflect on its confidence after providing an answer. In this work, we interpret an unsure confidence as an act of abstention. 

\subsection{Rewards Function Design}\label{sec:reward_func_design}
Figure~\ref{method}~(b) illustrates the proposed \Ours. We first partition the responses using a proxy model that captures semantic equivalence, then introduce the confidence reward and auxiliary rewards.
\subsubsection{Fine-grained Semantic Confidence Reward}
\textbf{Semantic Clustering.} To compute the semantic confidence reward for individual rollouts during GRPO training, we initially organize responses that express equivalent meanings through semantic clustering~\citep{SE}. These semantic clusters represent equivalence classes derived from a semantic equivalence relation that satisfies reflexivity, symmetry, and transitivity properties while capturing the semantic similarity between textual inputs. The practical implementation of equivalence via bidirectional entailment predictions generated by a Natural Language Inference (NLI) model, specifically DeBERTa~\citep{he2021deberta}, which determines the relationships between text pairs as either entailment, neutral, or contradiction. Texts are considered semantically equivalent when they mutually entail one another in both directions. We follow~\citet{SE} employing a greedy algorithm that assigns each response to an existing semantic cluster if it demonstrates semantic equivalence with any member of that cluster.

\textbf{Semantic Confidence}. The formulation of \Ours is based on the intuition that generative semantic consensus reflects model confidence. We assume that a larger semantic cluster implies strong consensus among the generated responses, reflecting higher confidence. Conversely, smaller or singleton clusters suggest divergent or less-supported meanings, thus implying lower confidence. Following this principle, we quantify the confidence of each rollout using a measure derived from the cardinality of the semantic cluster it belongs to. This per-sample approach distinguishes our method from previous work that relies on a single, global metric like semantic entropy calculated over all responses. For a given rollout $y_i$ and its corresponding semantic cluster $C_i$, the confidence reward function is defined as:
\begin{equation}
    R_{c} = 
    \begin{cases} 
    1, & \text{if ($\lvert C_i \rvert \geq \tau$ and $U = \text{``sure''}$) or ($\lvert C_i \rvert < \tau $ and $U = \text{``unsure''}$)} \\
    0, & \text{otherwise}
    \end{cases}
    \label{eq:uncertainty_reward}
\end{equation}
Here, $\tau$ is an empirical threshold of abstention, which we set to $ \lceil \frac{G}{2} \rceil$ in all experiments, where $G$ is the total number of rollouts. $U$ is the model's expressed confidence, either ``sure'' or ``unsure'', extracted from the \verb|<confidence> </confidence>| tag within the rollout. By incentivizing this alignment, the reward function effectively teaches the model a reliable abstention policy: to answer when semantic confidence is high and to abstain otherwise.

\subsubsection{Auxiliary Rewards}
\textbf{Accuracy Reward.} To prevent the model from engaging in reward hacking by learning to maximize the confidence reward on a uniform but incorrect answer, we introduce an auxiliary accuracy reward. This reward is granted only when the answer is semantically equivalent to the ground truth solution.
\begin{equation}
    R_a = 
    \begin{cases} 
    1, & \text{if the answer and the solution are semantically equivalent} \\
    0, & \text{otherwise}
    \end{cases}
    \label{eq:acc_reward_eq}
\end{equation}
where the answer is extracted between  \verb|<answer> </answer>| tags.

\textbf{Format Reward.} We encourage the model to enclose the answer within \verb|<answer> </answer>| tags and to present the confidence enclosed within \verb|<confidence> </confidence>| tags. The format reward $R_f$ is composed of two components. For each correctly used tag (\verb|<answer>|, \verb|</answer>|, \verb|<confidence>|, \verb|</confidence>|) that appears in the rollout, we assign a reward of $0.125$. To allow the model to determine its confidence based on its answer, it is necessary to place the answer before the confidence expression. We assign an additional reward of $0.5$ if the tags appear in the correct order. 

The total reward $R_{total}$ is defined as a weighted sum of three distinct components:
\begin{equation}
R_{total} = w_cR_c+w_aR_a+w_fR_f,
\label{eq:sum_reward}
\end{equation}
where $w_c$, $w_a$, $w_f$ are the respective weights for each reward component. Notably, only when the format is correct ($R_f=1$) will the other two rewards be added.

\section{Experiments}
\subsection{Experiment Setting}
\textbf{Datasets.} We use four datasets of question-answering (QA) task for evaluation, including \textbf{Pararel}~\citep{pararel_dataset} which contains cloze-style questions of relation prediction, \textbf{TriviaQA}~\citep{joshi-etal-2017-triviaqa} which contains general knowledge QA pairs, \textbf{Natural Questions (NQ)}~\citep{kwiatkowski-etal-2019-natural} which contains questions from aggregated queries to Google Search, and \textbf{SciQ}~\citep{welbl-etal-2017-crowdsourcing} which contains science exam questions. We use the \textbf{Pararel} dataset for training.

\textbf{Baselines.} We compare \Ours with three types of baselines.

\begin{enumerate}[leftmargin=1.2em]
    \item \textbf{Prompting method}: In-Context Learning (ICL), In-Context Learning with refusal examples (ICL-IDK), and In-Context Learning with confidence expression (ICL-Unsure). We take the performance of ICL as the performance of the initial model.
    \item \textbf{Supervised fine-tuning methods}: R-Tuning~\citep{R-tuning}, R-Tuning-U~\citep{R-tuning}, and SE-Tuning~\citep{SE_Tuning}.
    
    \item \textbf{RL-based method}: GRPO with semantic entropy (GRPO-SE), which adopts the training data as SE-Tuning and is built on GRPO.
\end{enumerate}

\textbf{Evaluation.} We use four metrics: $\text{F1}_{ans}$, $\text{F1}_{abs}$, $\text{F1}_{rel}$, and accuracy (Acc) for evaluation, which are detailed in Section~\ref{rel_metric_analyze}. We evaluate the correctness of the model's response by prompting with 6-shot examples. Following \citet{ualign}, we adopt bidirectional string matching to evaluate answer correctness. Here, we treat a response as an abstention if the confidence part is unsure. More details about the evaluation are listed in~\Cref{Evaluation_Prompt}. 

\textbf{Implementation Details.} We choose Llama3-8B-Instruct~\citep{llama3} and Qwen2.5-7B-Instruct~\citep{qwen2.5} in our experiments. The reinforcement learning framework is built on Open-R1~\citep{openr1}. We use Deberta-v2-xlarge-mnli~\citep{he2021deberta} as the NLI model for semantic clustering. The sampling number is set to 10, and the threshold $\tau$ is set to 5. All experiments are implemented on Nvidia L40-48GB GPUs. The training epoch is set to 1. We set the sampling temperature to 1.0 during the GRPO training. During inference, we utilize the vLLM framework to accelerate the process and employ a greedy search strategy to generate responses. Hyperparameters and additional configurations are detailed in~\Cref{Hyperparameter_Setup}.

\subsection{Main Results}

\begin{table}[t]
\small

\renewcommand{\arraystretch}{1.1}

\setlength{\tabcolsep}{2.3pt}
\caption{Performance on in-domain and out-of-domain knowledge-intensive QA datasets. All results are multiplied by 100. Notably, ICL is the initial model, so we do not calculate its F1 scores.}
\begin{center}
\begin{tabular}{l|llll|llll|llll|llll} 
\toprule
\multicolumn{1}{c}{\multirow{2}{*}{\begin{tabular}[c]{@{}c@{}}\textbf{Method}\\\end{tabular}}} & \multicolumn{4}{c}{\textbf{Pararel (ID)}}                               & \multicolumn{4}{c}{\textbf{TriviaQA (OOD)}}                             & \multicolumn{4}{c}{\textbf{NQ (OOD)}}                             & \multicolumn{4}{c}{\textbf{SciQ (OOD)}}                        \\ 
\cmidrule{2-17}
\multicolumn{1}{c}{} & $\text{F1}_{\text{ans}}$ & $\text{F1}_{\text{abs}}$ & $\text{F1}_{\text{rel}}$ & Acc & $\text{F1}_{\text{ans}}$ & $\text{F1}_{\text{abs}}$ & $\text{F1}_{\text{rel}}$ & Acc & $\text{F1}_{\text{ans}}$ & $\text{F1}_{\text{abs}}$ & $\text{F1}_{\text{rel}}$ & Acc & $\text{F1}_{\text{ans}}$ & $\text{F1}_{\text{abs}}$ & $\text{F1}_{\text{rel}}$ & Acc. \\

\midrule
\multicolumn{17}{c}{\textbf{Llama3-8B-Instruct}}                                                                                                                                                                                                                                                                                                                                              \\ 
\midrule
ICL                                                                                            & -             & -             & -             & 48.8                    & -             & -             & -             & \textbf{68.4}           & -             & -             & -             & \textbf{37.1}           & -             & -             & -             & \textbf{65.3}  \\
ICL-IDK                                                                                        & 62.1          & 64.3          & 63.2          & 31.7                    & 69.8          & 32.6          & 44.5          & 55.3                    & 43.5          & 67.9          & 53.0          & 18.1                    & 71.8          & 43.9          & 54.5          & 56.4           \\
ICL-Unsure                                                                                     & 70.4          & 68.2          & 69.3          & 41.9                    & 67.7          & 46.4          & 55.0          & 51.7                    & 45.9          & 18.0          & 25.9          & 30.4                    & 64.7          & 17.9          & 28.0          & 53.1           \\
R-Tuning                                                                                       & 78.3          & \textbf{79.6} & \textbf{79.0} & 40.5                    & 47.7          & 51.5          & 49.5          & 26.3                    & 31.2          & 72.0          & 43.5          & 10.8                    & 32.1          & 51.4          & 39.5          & 14.6           \\
R-Tuning-U                                                                                     & 73.8          & 76.6          & 75.2          & 40.5                    & 42.6          & 53.2          & 47.3          & 20.4                    & 31.8          & \textbf{74.0}          & 44.5          & 10.3                    & 49.9          & 49.7          & 49.8          & 28.0           \\
SE-Tuning   & 78.2          & 75.0          & 76.6          & 45.9                    & 44.9          & 50.7          & 47.7          & 24.0                    & 27.7          & 72.2          & 40.0          & 9.4                     & 22.4          & 50.4          & 31.0          & 9.2            \\
GRPO-SE                                                                                        & 70.9          & 77.3          & 73.9          & 35.1                    & 70.5          & 60.2          & 65.0          & 47.3                    & 37.2          & 73.9          & 49.5          & 13.0                    & 64.4          & 51.2          & 57.0          & 46.4           \\
\Ours    & \textbf{83.3} & 67.1          & 74.3          & \textbf{59.4}           & \textbf{82.6} & \textbf{62.0} & \textbf{70.8} & 61.7                    & \textbf{53.1}  & 72.5          & \textbf{61.3} & 21.7  & \textbf{74.9} & \textbf{53.3} & \textbf{62.3} & 52.4           \\ 
\midrule
\multicolumn{17}{c}{\textbf{Qwen2.5-7B-Instruct}}                                                                                                                                                                                                                                                                                                                                             \\ 
\midrule
ICL                                                                                            & -             & -             & -             & 44.1                    & -             & -             & -             & \textbf{51.7}           & -             & -             & -             & \textbf{27.6}           & -             & -             & -             & \textbf{68.8}  \\
ICL-IDK                                                                                        & 60.1          & 68.6          & 64.1          & 27.4                    & 66.6          & 49.3          & 56.6          & 44.9                    & 38.1          & 50.1          & 43.3          & 19.4                    & 71.6          & 14.8          & 24.5          & 60.5           \\
ICL-Unsure                                                                                     & 77.7          & 82.9          & 80.2          & 35.8                    & 71.2          & 51.7          & 59.9          & 49.8                    & 40.2          & 39.7          & 39.9          & 21.9                    & 71.4          & 37.6          & 49.2          & 57.3           \\
R-Tuning                                                                                       & 79.3          & \textbf{86.4} & \textbf{82.7} & 31.8                    & 44.0          & 71.0          & 54.4          & 15.1                    & 25.1          & 84.5 & 38.7          & 4.5                     & 43.8          & 53.9          & 48.3          & 19.9           \\
R-Tuning-U                                                                                     & 62.1          & 80.5          & 70.1          & 21.3                    & 40.1          & 69.8          & 51.0          & 13.6                     & 32.3          & \textbf{84.8}          & 46.7          & 6.3                     & 43.9           & 52.5          & 47.8           & 20.4     \\
SE-Tuning       & 71.4          & 82.3          & 76.5          & 28.0                    & 33.8          & 69.3          & 45.4          & 10.7                    & 32.8          & 84.2          & 47.2          & 6.7                     & 55.4          & 52.7          & 54.0         & 34.5           \\
GRPO-SE                                                                                        & 75.2          & 83.3          & 79.0          & 31.7                    & 71.6          & 72.4          & 72.0          & 41.6                    & 42.8          & 78.0          & 55.3          & 14.2                    & 76.7          & 50.7          & 61.0          & 57.3           \\
\Ours                                                                                           & \textbf{80.8} & 72.9          & 76.6          & \textbf{51.9}           & \textbf{83.3} & \textbf{79.1} & \textbf{81.1} & 51.3                    & \textbf{57.3} & 80.1          & \textbf{66.8} & 20.4                    & \textbf{79.2} & \textbf{54.5} & \textbf{64.6} & 58.0           \\
\bottomrule
\end{tabular}
\end{center}
\label{main_result}
\end{table}

We present the main experimental results of \Ours and all baseline methods in Table~\ref{main_result}, and we show the methods based on different LLMs. The results reveal several key findings:

\textbf{Generalizable Reliability.} We focus on metric $\text{F1}_{rel}$ that assesses a model's overall reliability by balancing helpfulness ($\text{F1}_{ans}$) and truthfulness ($\text{F1}_{abs}$). A key finding is that the high reliability demonstrated by some baselines on in-domain (ID) data is brittle and does not generalize. For example, R-Tuning using the Llama model achieves a high $\text{F1}_{rel}$ score of 79.0 on the Pararel (ID) dataset by directly learning to abstain from unknown questions. However, this specialized performance collapses on out-of-distribution (OOD) tasks, with its score dropping to just 49.5 of $\text{F1}_{rel}$ in TriviaQA. In contrast, the proposed \Ours demonstrates a more robust, reliable performance that excels at generalization. It achieves this by attaining a superior balance between answering known questions ($\text{F1}_{ans}$ of 83.3) and correctly abstaining ($\text{F1}_{abs}$ of 67.1). This balanced approach leads to a high $\text{F1}_{rel}$ score of 74.3 on the ID dataset, which shows a much smaller drop to 70.8 on the OOD TriviaQA dataset. This consistent outperformance on all three OOD datasets proves that the knowledge boundary learned by our approach is more generalizable. Compared to SE-Tuning, GRPO-SE shows stronger generalization: SE-Tuning performs well on ID data but degrades significantly on OOD sets, while GRPO-SE maintains robust performance across both, highlighting the advantage of reinforcement learning over supervised fine-tuning. 

\textbf{Effectiveness of the Fine-grained Semantic Confidence Reward.} We compare our approach directly with GRPO-SE, a baseline employing the same GRPO algorithm but utilizing a coarse semantic uncertainty reward. The results demonstrate a clear advantage: our method \Ours significantly outperforms GRPO-SE across all OOD datasets and different LLMs in terms of $\text{F1}_{rel}$. Given that the primary distinction between the two methods is the reward function, this comparison provides empirical evidence that our fine-grained confidence reward is substantially more effective than a coarse, global entropy-based signal. 

\textbf{Comparison between Different LLMs.} We fine-tuned both the Llama3-8B-Instruct and Qwen2.5-7B-Instruct models on the Pararel dataset using our \Ours method alongside established baseline abstention techniques. From the results on TriviaQA and NQ we can see that although Llama achieved higher raw accuracy than Qwen both before and after fine-tuning, its overall reliability (measured by $\text{F1}_{rel}$) remained lower. This disparity indicates that a model's extensive knowledge capacity does not ensure precise self-awareness of its knowledge boundaries. It also highlights that accuracy alone provides an inadequate proxy for gauging improvements in abstention alignment. 
\subsection{Ablation Study}
\begin{figure}[htbp]
\begin{center}
\includegraphics[width=0.85\linewidth]{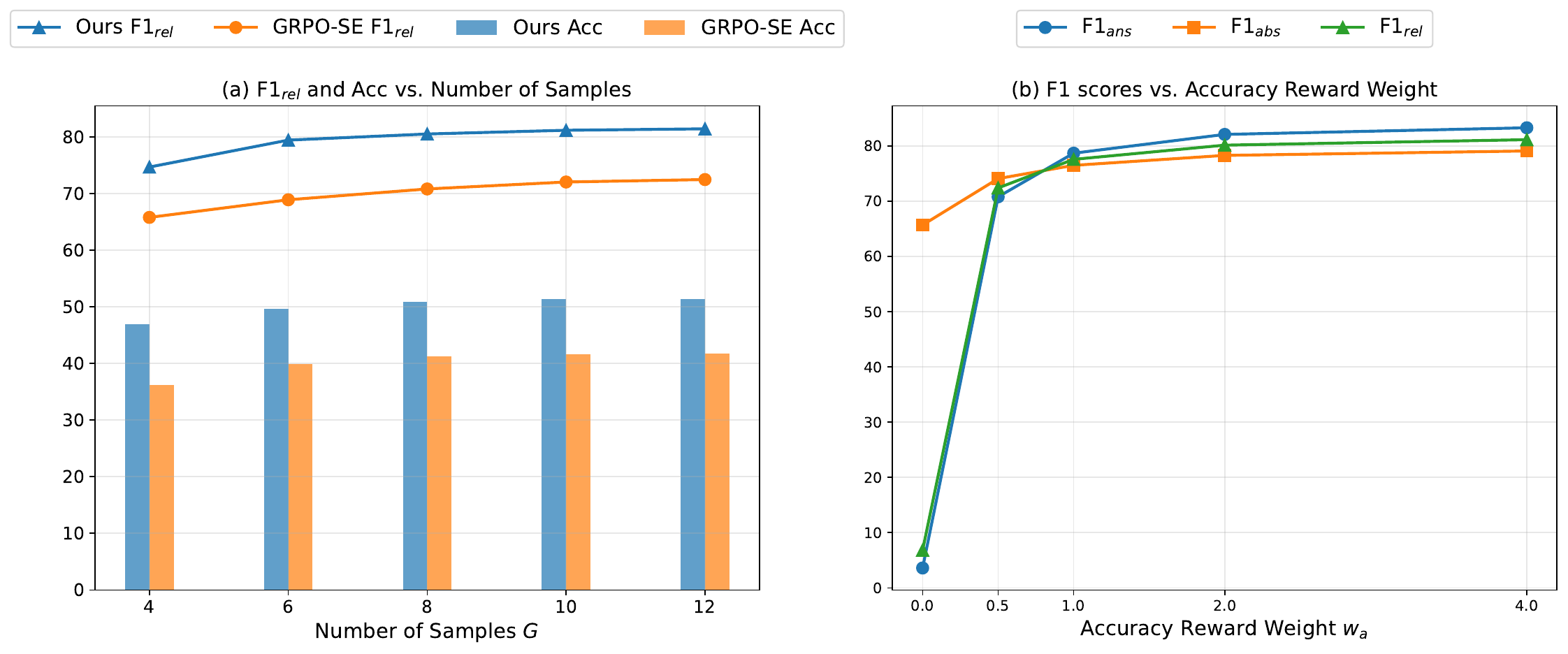}
\end{center}
\caption{(a) Experiments of $\text{F1}_{rel}$ and accuracy of various sampling number $G$ on TriviaQA on Qwen2.5-7B-Instruct. (b) Experiments of $\text{F1}_{ans}$, $\text{F1}_{abs}$ and $\text{F1}_{rel}$ of various accuracy reward weight $w_a$ on TriviaQA on Qwen2.5-7B-Instruct.}
\label{ablation_results}
\end{figure}
\textbf{Impact of Sampling Number.} We investigate the impact of the sampling number $G$ on performance in Figure~\ref{ablation_results}~(a). The results demonstrate that for our method \Ours, both reliability ($\text{F1}_{rel}$) and accuracy improve as $G$ increases, consistently outperforming the GRPO-SE baseline. Both metrics reach their optimal or near-optimal levels at $G=10$. Since performance gains become negligible beyond this point, we select $G=10$ as the default setting for all experiments to maximize performance while maintaining computational efficiency.

\textbf{Impact of Accuracy Reward Weight.} We evaluate the impact of the accuracy reward weight $w_a$ while fixing the confidence reward weight $w_c$ at 1. When $w_a$ is set to 0, the model achieves a high $\text{F1}_{abs}$ but a near-zero $\text{F1}_{ans}$, indicating that the model abstains from almost all questions. This occurs because the model receives rewards for generating low-confidence incorrect answers, a phenomenon known as reward hacking. Introducing the accuracy reward improves $\text{F1}_{rel}$. As the accuracy reward weight increases, both $\text{F1}_{ans}$ and $\text{F1}_{rel}$ rise steadily. Notably, weights greater than 1 produce higher $\text{F1}_{rel}$ scores, likely because obtaining accuracy rewards is more difficult than obtaining confidence rewards, thus requiring greater emphasis.

\subsection{Detailed Analysis}
\begin{figure}[!ht]
    \centering
    \includegraphics[width=0.8\linewidth]{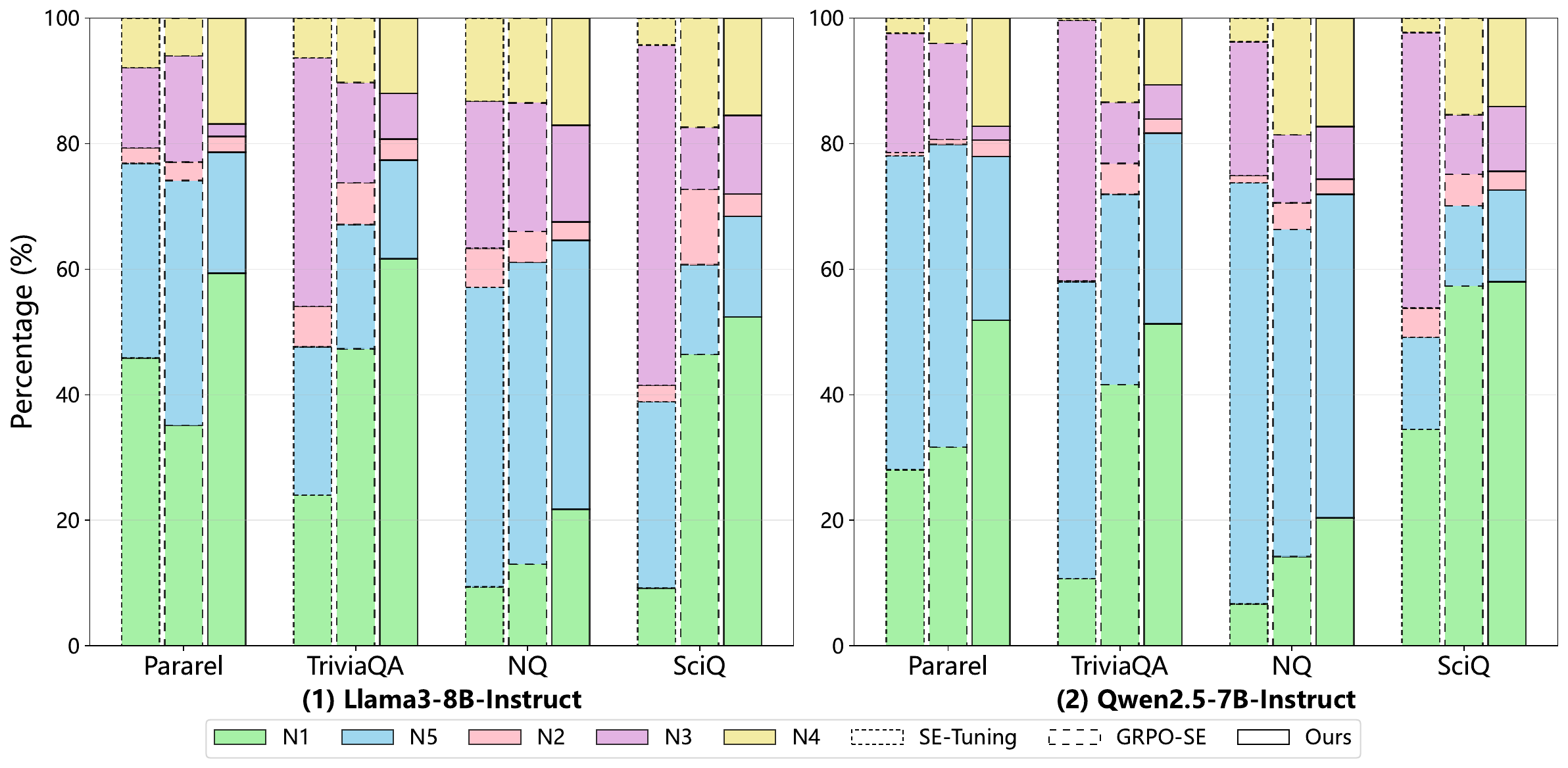}
    \caption{Percentage of prediction types among different methods. We choose SE-Tuning, GRPO-SE, and \Ours.}
    \label{fig:detail_analysis}
\end{figure}
Figure~\ref{fig:detail_analysis} compares the performance of three methods: SE-Tuning, GRPO-SE, and \Ours. The primary goal is to maximize correctly answered known questions ($N_1$) and abstained unknown questions ($N_5$), while minimizing errors ($N_2$, $N_4$) and unnecessary abstentions ($N_3$). Note that Pararel is the ID dataset, whereas all the others are OOD. On OOD datasets, SE-Tuning tends to over-abstain, reflected by a larger $N_3$, indicating a conservative strategy that suppresses hallucinations at the cost of unnecessary refusals on answerable cases. By contrast, GRPO-SE and \Ours show higher hallucination rates than SE-Tuning, which reflects the seesaw effect between helpfulness and truthfulness: reducing over-abstention encourages the model to answer more often and thus increases the chance of hallucinations relative to an overly conservative baseline. Importantly, in most cases \Ours attains a higher $N_1{+}N_5$ than the baselines, indicating a better overall balance between answering known questions and abstaining on unknown ones, and hence higher reliability.

\section{Related Work}
\label{related_work}
\textbf{Uncertainty Estimation and Confidence Elicitation.} Uncertainty estimation aims to quantify this risk before LLMs respond to a request, which is an effective tool for detecting and mitigating hallucinations of LLMs~\citep{UQ_survey}. Likelihood-based methods estimate uncertainty by computing probabilities for salient tokens, capturing the model’s confidence in specific outputs~\citep{duan-etal-2024-shifting, lin-etal-2024-contextualized}. Sampling-based methods generate multiple candidate responses, cluster them, and quantify uncertainty based on the degree of inter-response consistency~\citep{SE, semantic_density, nikitin2024kernel, aichberger2025improving}. Probing-based methods leverage internal model representations to train classifiers that predict uncertainty or detect hallucinations, often without requiring additional generation \citep{SEP,geometry_of_truth, estimating_without_generate, probing_llm_lying}. Confidence elicitation aims to elicit the confidence of LLMs in their answers~\citep{confidence_estimation-survey}. Prompting methods directly prompt LLMs to express confidence~\citep{lin2022teaching, kadavath2022language}. Recent studies~\citep{xu-etal-2024-sayself,damani2025beyond} employ chain-of-thought (CoT) training to teach LLMs to reason about their confidence. In this work, we adopt binary confidence elicitation (``sure'' vs. ``unsure'') as a proxy for abstention, enabling models to discard low-confidence answers post hoc.

\textbf{Abstention Fine-Tuning.} Many studies explore fine-tuning LLMs to abstain from answering questions beyond their knowledge boundaries~\citep{abstention_survey}. \citet{brahman2024the,cheng2024can,xurejection} propose learning from preferences via direct preference optimization~\citep{DPO} to train LLMs to abstain from answering unknown questions. \citet{li2025refine} introduce adaptive contrastive preference learning to calibrate abstention behavior. \citet{R-tuning, SE_Tuning,ualign} leverage uncertainty to construct abstention-aware datasets, replacing the ground-truth labels of uncertain questions with ``I don't know''. \citet{yang2025barrel,deng-etal-2024-dont} train LLMs to reason about their own uncertainty and provide post-hoc explanations for why a given question is unanswerable. \citet{rej_token,idk_token} introduce a special ``rejection'' token into the model’s vocabulary and design an objective function that reallocates probability mass toward the ``rejection'' token when the model encounters uncertain predictions. In this work, we go beyond aggregated uncertainty by utilizing fine-grained, per-sample confidence to sharpen the model’s awareness of its knowledge limits.

\section{Conclusion}
In this paper, we focus on enhancing the reliability of LLMs, which we define as the ability to accurately answer questions within their knowledge boundaries and abstain from those beyond them. First, we propose a new comprehensive reliability metric by calculating the harmonic mean of the F1 scores for answering and abstaining. This metric integrates model helpfulness and truthfulness into a single, robust score that is monotonically sensitive to all error categories. We then introduce \Ours, a reinforcement learning method with a fine-grained semantic confidence reward. This reward mechanism, derived from the semantic confidence among multiple generated responses, guides the model to retain the answer only when generative consensus is high and to discard it when it is low, thereby enabling the model to make abstention decisions more aligned with its knowledge boundaries. Our experiments on both in-domain and out-of-distribution datasets show that \Ours achieves a better balance between helpfulness and truthfulness than baselines, thereby improving overall model reliability.

\bibliography{iclr2026_conference}

\begin{thebibliography}{47}
\providecommand{\natexlab}[1]{#1}
\providecommand{\url}[1]{\texttt{#1}}
\expandafter\ifx\csname urlstyle\endcsname\relax
  \providecommand{\doi}[1]{doi: #1}\else
  \providecommand{\doi}{doi: \begingroup \urlstyle{rm}\Url}\fi

\bibitem[Achiam et~al.(2023)Achiam, Adler, Agarwal, Ahmad, Akkaya, Aleman, Almeida, Altenschmidt, Altman, Anadkat, et~al.]{gpt4}
Josh Achiam, Steven Adler, Sandhini Agarwal, Lama Ahmad, Ilge Akkaya, Florencia~Leoni Aleman, Diogo Almeida, Janko Altenschmidt, Sam Altman, Shyamal Anadkat, et~al.
\newblock Gpt-4 technical report.
\newblock \emph{arXiv preprint arXiv:2303.08774}, 2023.

\bibitem[Aichberger et~al.(2025)Aichberger, Schweighofer, Ielanskyi, and Hochreiter]{aichberger2025improving}
Lukas Aichberger, Kajetan Schweighofer, Mykyta Ielanskyi, and Sepp Hochreiter.
\newblock Improving uncertainty estimation through semantically diverse language generation.
\newblock In \emph{The Thirteenth International Conference on Learning Representations}, 2025.
\newblock URL \url{https://openreview.net/forum?id=HSi4VetQLj}.

\bibitem[Azaria \& Mitchell(2023)Azaria and Mitchell]{probing_llm_lying}
Amos Azaria and Tom Mitchell.
\newblock The internal state of an {LLM} knows when it's lying.
\newblock In \emph{The 2023 Conference on Empirical Methods in Natural Language Processing}, 2023.
\newblock URL \url{https://openreview.net/forum?id=y2V6YgLaW7}.

\bibitem[Brahman et~al.(2024)Brahman, Kumar, Balachandran, Dasigi, Pyatkin, Ravichander, Wiegreffe, Dziri, Chandu, Hessel, Tsvetkov, Smith, Choi, and Hajishirzi]{brahman2024the}
Faeze Brahman, Sachin Kumar, Vidhisha Balachandran, Pradeep Dasigi, Valentina Pyatkin, Abhilasha Ravichander, Sarah Wiegreffe, Nouha Dziri, Khyathi Chandu, Jack Hessel, Yulia Tsvetkov, Noah~A. Smith, Yejin Choi, and Hannaneh Hajishirzi.
\newblock The art of saying no: Contextual noncompliance in language models.
\newblock In \emph{The Thirty-eight Conference on Neural Information Processing Systems Datasets and Benchmarks Track}, 2024.
\newblock URL \url{https://openreview.net/forum?id=f1UL4wNlw6}.

\bibitem[Cheng et~al.(2024)Cheng, Sun, Liu, Zhang, Yin, Li, Li, He, Chen, and Qiu]{cheng2024can}
Qinyuan Cheng, Tianxiang Sun, Xiangyang Liu, Wenwei Zhang, Zhangyue Yin, Shimin Li, Linyang Li, Zhengfu He, Kai Chen, and Xipeng Qiu.
\newblock Can {AI} assistants know what they don't know?
\newblock In \emph{Forty-first International Conference on Machine Learning}, 2024.
\newblock URL \url{https://openreview.net/forum?id=girxGkdECL}.

\bibitem[Chu et~al.(2025)Chu, Zhai, Yang, Tong, Xie, Schuurmans, Le, Levine, and Ma]{sft_vs_rl}
Tianzhe Chu, Yuexiang Zhai, Jihan Yang, Shengbang Tong, Saining Xie, Dale Schuurmans, Quoc~V. Le, Sergey Levine, and Yi~Ma.
\newblock Sft memorizes, rl generalizes: A comparative study of foundation model post-training, 2025.
\newblock URL \url{https://arxiv.org/abs/2501.17161}.

\bibitem[Cohen et~al.(2024)Cohen, Dobler, Biran, and de~Melo]{idk_token}
Roi Cohen, Konstantin Dobler, Eden Biran, and Gerard de~Melo.
\newblock I don\textquotesingle t know: Explicit modeling of uncertainty with an [idk] token.
\newblock In A.~Globerson, L.~Mackey, D.~Belgrave, A.~Fan, U.~Paquet, J.~Tomczak, and C.~Zhang (eds.), \emph{Advances in Neural Information Processing Systems}, volume~37, pp.\  10935--10958. Curran Associates, Inc., 2024.
\newblock URL \url{https://proceedings.neurips.cc/paper_files/paper/2024/file/14c018d2e72c521605b0567029ef0efb-Paper-Conference.pdf}.

\bibitem[Damani et~al.(2025)Damani, Puri, Slocum, Shenfeld, Choshen, Kim, and Andreas]{damani2025beyond}
Mehul Damani, Isha Puri, Stewart Slocum, Idan Shenfeld, Leshem Choshen, Yoon Kim, and Jacob Andreas.
\newblock Beyond binary rewards: Training lms to reason about their uncertainty.
\newblock \emph{arXiv preprint arXiv:2507.16806}, 2025.

\bibitem[Deng et~al.(2024)Deng, Zhao, Li, Ng, and Chua]{deng-etal-2024-dont}
Yang Deng, Yong Zhao, Moxin Li, See-Kiong Ng, and Tat-Seng Chua.
\newblock Don{'}t just say ``{I} don{'}t know''! self-aligning large language models for responding to unknown questions with explanations.
\newblock In Yaser Al-Onaizan, Mohit Bansal, and Yun-Nung Chen (eds.), \emph{Proceedings of the 2024 Conference on Empirical Methods in Natural Language Processing}, pp.\  13652--13673, Miami, Florida, USA, November 2024. Association for Computational Linguistics.
\newblock \doi{10.18653/v1/2024.emnlp-main.757}.
\newblock URL \url{https://aclanthology.org/2024.emnlp-main.757/}.

\bibitem[Duan et~al.(2024)Duan, Cheng, Wang, Zavalny, Wang, Xu, Kailkhura, and Xu]{duan-etal-2024-shifting}
Jinhao Duan, Hao Cheng, Shiqi Wang, Alex Zavalny, Chenan Wang, Renjing Xu, Bhavya Kailkhura, and Kaidi Xu.
\newblock Shifting attention to relevance: Towards the predictive uncertainty quantification of free-form large language models.
\newblock In Lun-Wei Ku, Andre Martins, and Vivek Srikumar (eds.), \emph{Proceedings of the 62nd Annual Meeting of the Association for Computational Linguistics (Volume 1: Long Papers)}, pp.\  5050--5063, Bangkok, Thailand, August 2024. Association for Computational Linguistics.
\newblock \doi{10.18653/v1/2024.acl-long.276}.
\newblock URL \url{https://aclanthology.org/2024.acl-long.276/}.

\bibitem[Elazar et~al.(2021)Elazar, Kassner, Ravfogel, Ravichander, Hovy, Sch{\"u}tze, and Goldberg]{pararel_dataset}
Yanai Elazar, Nora Kassner, Shauli Ravfogel, Abhilasha Ravichander, Eduard Hovy, Hinrich Sch{\"u}tze, and Yoav Goldberg.
\newblock Measuring and improving consistency in pretrained language models.
\newblock \emph{Transactions of the Association for Computational Linguistics}, 9:\penalty0 1012--1031, 2021.

\bibitem[Face(2025)]{openr1}
Hugging Face.
\newblock Open r1: A fully open reproduction of deepseek-r1, January 2025.
\newblock URL \url{https://github.com/huggingface/open-r1}.

\bibitem[Feng et~al.(2024)Feng, Shi, Wang, Ding, Balachandran, and Tsvetkov]{feng-etal-2024-dont}
Shangbin Feng, Weijia Shi, Yike Wang, Wenxuan Ding, Vidhisha Balachandran, and Yulia Tsvetkov.
\newblock Don{'}t hallucinate, abstain: Identifying {LLM} knowledge gaps via multi-{LLM} collaboration.
\newblock In Lun-Wei Ku, Andre Martins, and Vivek Srikumar (eds.), \emph{Proceedings of the 62nd Annual Meeting of the Association for Computational Linguistics (Volume 1: Long Papers)}, pp.\  14664--14690, Bangkok, Thailand, August 2024. Association for Computational Linguistics.
\newblock \doi{10.18653/v1/2024.acl-long.786}.
\newblock URL \url{https://aclanthology.org/2024.acl-long.786/}.

\bibitem[Geng et~al.(2024)Geng, Cai, Wang, Koeppl, Nakov, and Gurevych]{confidence_estimation-survey}
Jiahui Geng, Fengyu Cai, Yuxia Wang, Heinz Koeppl, Preslav Nakov, and Iryna Gurevych.
\newblock A survey of confidence estimation and calibration in large language models.
\newblock In Kevin Duh, Helena Gomez, and Steven Bethard (eds.), \emph{Proceedings of the 2024 Conference of the North American Chapter of the Association for Computational Linguistics: Human Language Technologies (Volume 1: Long Papers)}, pp.\  6577--6595, Mexico City, Mexico, June 2024. Association for Computational Linguistics.
\newblock \doi{10.18653/v1/2024.naacl-long.366}.
\newblock URL \url{https://aclanthology.org/2024.naacl-long.366/}.

\bibitem[Gottesman \& Geva(2024)Gottesman and Geva]{estimating_without_generate}
Daniela Gottesman and Mor Geva.
\newblock Estimating knowledge in large language models without generating a single token.
\newblock In Yaser Al-Onaizan, Mohit Bansal, and Yun-Nung Chen (eds.), \emph{Proceedings of the 2024 Conference on Empirical Methods in Natural Language Processing}, pp.\  3994--4019, Miami, Florida, USA, November 2024. Association for Computational Linguistics.
\newblock \doi{10.18653/v1/2024.emnlp-main.232}.
\newblock URL \url{https://aclanthology.org/2024.emnlp-main.232/}.

\bibitem[Grattafiori et~al.(2024)Grattafiori, Dubey, Jauhri, Pandey, Kadian, Al-Dahle, Letman, Mathur, Schelten, Vaughan, et~al.]{llama3}
Aaron Grattafiori, Abhimanyu Dubey, Abhinav Jauhri, Abhinav Pandey, Abhishek Kadian, Ahmad Al-Dahle, Aiesha Letman, Akhil Mathur, Alan Schelten, Alex Vaughan, et~al.
\newblock The llama 3 herd of models.
\newblock \emph{arXiv preprint arXiv:2407.21783}, 2024.

\bibitem[Guo et~al.(2025)Guo, Yang, Zhang, Song, Zhang, Xu, Zhu, Ma, Wang, Bi, et~al.]{guo2025deepseek-r1}
Daya Guo, Dejian Yang, Haowei Zhang, Junxiao Song, Ruoyu Zhang, Runxin Xu, Qihao Zhu, Shirong Ma, Peiyi Wang, Xiao Bi, et~al.
\newblock Deepseek-r1: Incentivizing reasoning capability in llms via reinforcement learning.
\newblock \emph{arXiv preprint arXiv:2501.12948}, 2025.

\bibitem[Han et~al.(2024)Han, Kossen, Razzak, Schut, Malik, and Gal]{SEP}
Jiatong Han, Jannik Kossen, Muhammed Razzak, Lisa Schut, Shreshth~A Malik, and Yarin Gal.
\newblock Semantic entropy probes: Robust and cheap hallucination detection in {LLM}s.
\newblock In \emph{ICML 2024 Workshop on Foundation Models in the Wild}, 2024.
\newblock URL \url{https://openreview.net/forum?id=Zd0XLr6JKn}.

\bibitem[He et~al.(2021)He, Liu, Gao, and Chen]{he2021deberta}
Pengcheng He, Xiaodong Liu, Jianfeng Gao, and Weizhu Chen.
\newblock Deberta: Decoding-enhanced bert with disentangled attention.
\newblock In \emph{International Conference on Learning Representations}, 2021.
\newblock URL \url{https://openreview.net/forum?id=XPZIaotutsD}.

\bibitem[Huang et~al.(2025)Huang, Feng, Ma, Fan, Feng, Gu, Ye, Zhao, Zhong, Wang, Wu, Hu, Kong, Xiao, Liu, and Qin]{rej_token}
Lei Huang, Xiaocheng Feng, Weitao Ma, Yuchun Fan, Xiachong Feng, Yuxuan Gu, Yangfan Ye, Liang Zhao, Weihong Zhong, Baoxin Wang, Dayong Wu, Guoping Hu, Lingpeng Kong, Tong Xiao, Ting Liu, and Bing Qin.
\newblock Alleviating hallucinations from knowledge misalignment in large language models via selective abstention learning.
\newblock In Wanxiang Che, Joyce Nabende, Ekaterina Shutova, and Mohammad~Taher Pilehvar (eds.), \emph{Proceedings of the 63rd Annual Meeting of the Association for Computational Linguistics (Volume 1: Long Papers)}, pp.\  24564--24579, Vienna, Austria, July 2025. Association for Computational Linguistics.
\newblock ISBN 979-8-89176-251-0.
\newblock \doi{10.18653/v1/2025.acl-long.1199}.
\newblock URL \url{https://aclanthology.org/2025.acl-long.1199/}.

\bibitem[Joshi et~al.(2017)Joshi, Choi, Weld, and Zettlemoyer]{joshi-etal-2017-triviaqa}
Mandar Joshi, Eunsol Choi, Daniel Weld, and Luke Zettlemoyer.
\newblock {T}rivia{QA}: A large scale distantly supervised challenge dataset for reading comprehension.
\newblock In Regina Barzilay and Min-Yen Kan (eds.), \emph{Proceedings of the 55th Annual Meeting of the Association for Computational Linguistics (Volume 1: Long Papers)}, pp.\  1601--1611, Vancouver, Canada, July 2017. Association for Computational Linguistics.
\newblock \doi{10.18653/v1/P17-1147}.
\newblock URL \url{https://aclanthology.org/P17-1147/}.

\bibitem[Kadavath et~al.(2022)Kadavath, Conerly, Askell, Henighan, Drain, Perez, Schiefer, Hatfield-Dodds, DasSarma, Tran-Johnson, et~al.]{kadavath2022language}
Saurav Kadavath, Tom Conerly, Amanda Askell, Tom Henighan, Dawn Drain, Ethan Perez, Nicholas Schiefer, Zac Hatfield-Dodds, Nova DasSarma, Eli Tran-Johnson, et~al.
\newblock Language models (mostly) know what they know.
\newblock \emph{arXiv preprint arXiv:2207.05221}, 2022.

\bibitem[Kim et~al.(2024)Kim, Kim, Park, Kim, Park, Yoo, Lee, and Kim]{kim-etal-2024-aligning}
Hyuhng~Joon Kim, Youna Kim, Cheonbok Park, Junyeob Kim, Choonghyun Park, Kang~Min Yoo, Sang-goo Lee, and Taeuk Kim.
\newblock Aligning language models to explicitly handle ambiguity.
\newblock In Yaser Al-Onaizan, Mohit Bansal, and Yun-Nung Chen (eds.), \emph{Proceedings of the 2024 Conference on Empirical Methods in Natural Language Processing}, pp.\  1989--2007, Miami, Florida, USA, November 2024. Association for Computational Linguistics.
\newblock \doi{10.18653/v1/2024.emnlp-main.119}.
\newblock URL \url{https://aclanthology.org/2024.emnlp-main.119/}.

\bibitem[Kim et~al.(2025)Kim, Kim, Lee, and Kim]{when_to_speak}
Hyuhng~Joon Kim, Youna Kim, Sang-goo Lee, and Taeuk Kim.
\newblock When to speak, when to abstain: Contrastive decoding with abstention.
\newblock In Wanxiang Che, Joyce Nabende, Ekaterina Shutova, and Mohammad~Taher Pilehvar (eds.), \emph{Proceedings of the 63rd Annual Meeting of the Association for Computational Linguistics (Volume 1: Long Papers)}, pp.\  9710--9730, Vienna, Austria, July 2025. Association for Computational Linguistics.
\newblock ISBN 979-8-89176-251-0.
\newblock \doi{10.18653/v1/2025.acl-long.479}.
\newblock URL \url{https://aclanthology.org/2025.acl-long.479/}.

\bibitem[Kuhn et~al.(2023)Kuhn, Gal, and Farquhar]{SE}
Lorenz Kuhn, Yarin Gal, and Sebastian Farquhar.
\newblock Semantic uncertainty: Linguistic invariances for uncertainty estimation in natural language generation.
\newblock In \emph{The Eleventh International Conference on Learning Representations}, 2023.
\newblock URL \url{https://openreview.net/forum?id=VD-AYtP0dve}.

\bibitem[Kwiatkowski et~al.(2019)Kwiatkowski, Palomaki, Redfield, Collins, Parikh, Alberti, Epstein, Polosukhin, Devlin, Lee, Toutanova, Jones, Kelcey, Chang, Dai, Uszkoreit, Le, and Petrov]{kwiatkowski-etal-2019-natural}
Tom Kwiatkowski, Jennimaria Palomaki, Olivia Redfield, Michael Collins, Ankur Parikh, Chris Alberti, Danielle Epstein, Illia Polosukhin, Jacob Devlin, Kenton Lee, Kristina Toutanova, Llion Jones, Matthew Kelcey, Ming-Wei Chang, Andrew~M. Dai, Jakob Uszkoreit, Quoc Le, and Slav Petrov.
\newblock Natural questions: A benchmark for question answering research.
\newblock \emph{Transactions of the Association for Computational Linguistics}, 7:\penalty0 452--466, 2019.
\newblock \doi{10.1162/tacl_a_00276}.
\newblock URL \url{https://aclanthology.org/Q19-1026/}.

\bibitem[Li et~al.(2025{\natexlab{a}})Li, Zhao, Zhang, Li, Xie, Ng, Chua, and Deng]{li-etal-2025-knowledge-boundary}
Moxin Li, Yong Zhao, Wenxuan Zhang, Shuaiyi Li, Wenya Xie, See-Kiong Ng, Tat-Seng Chua, and Yang Deng.
\newblock Knowledge boundary of large language models: A survey.
\newblock In \emph{Proceedings of the 63rd Annual Meeting of the Association for Computational Linguistics (Volume 1: Long Papers)}, pp.\  5131--5157, Vienna, Austria, July 2025{\natexlab{a}}. Association for Computational Linguistics.
\newblock ISBN 979-8-89176-251-0.
\newblock URL \url{https://aclanthology.org/2025.acl-long.256/}.

\bibitem[Li et~al.(2025{\natexlab{b}})Li, Huang, Kuang, Li, Guo, Qu, Tan, Zheng, Shen, and Yu]{li2025refine}
Yinghui Li, Haojing Huang, Jiayi Kuang, Yangning Li, Shu-Yu Guo, Chao Qu, Xiaoyu Tan, Hai-Tao Zheng, Ying Shen, and Philip~S. Yu.
\newblock Refine knowledge of large language models via adaptive contrastive learning.
\newblock In \emph{The Thirteenth International Conference on Learning Representations}, 2025{\natexlab{b}}.
\newblock URL \url{https://openreview.net/forum?id=HqjRlT65WX}.

\bibitem[Lin et~al.(2022)Lin, Hilton, and Evans]{lin2022teaching}
Stephanie Lin, Jacob Hilton, and Owain Evans.
\newblock Teaching models to express their uncertainty in words.
\newblock \emph{Transactions on Machine Learning Research}, 2022.
\newblock ISSN 2835-8856.
\newblock URL \url{https://openreview.net/forum?id=8s8K2UZGTZ}.

\bibitem[Lin et~al.(2024)Lin, Trivedi, and Sun]{lin-etal-2024-contextualized}
Zhen Lin, Shubhendu Trivedi, and Jimeng Sun.
\newblock Contextualized sequence likelihood: Enhanced confidence scores for natural language generation.
\newblock In Yaser Al-Onaizan, Mohit Bansal, and Yun-Nung Chen (eds.), \emph{Proceedings of the 2024 Conference on Empirical Methods in Natural Language Processing}, pp.\  10351--10368, Miami, Florida, USA, November 2024. Association for Computational Linguistics.
\newblock \doi{10.18653/v1/2024.emnlp-main.578}.
\newblock URL \url{https://aclanthology.org/2024.emnlp-main.578/}.

\bibitem[Marks \& Tegmark(2024)Marks and Tegmark]{geometry_of_truth}
Samuel Marks and Max Tegmark.
\newblock The geometry of truth: Emergent linear structure in large language model representations of true/false datasets.
\newblock In \emph{First Conference on Language Modeling}, 2024.
\newblock URL \url{https://openreview.net/forum?id=aajyHYjjsk}.

\bibitem[Nikitin et~al.(2024)Nikitin, Kossen, Gal, and Marttinen]{nikitin2024kernel}
Alexander~V Nikitin, Jannik Kossen, Yarin Gal, and Pekka Marttinen.
\newblock Kernel language entropy: Fine-grained uncertainty quantification for {LLM}s from semantic similarities.
\newblock In \emph{The Thirty-eighth Annual Conference on Neural Information Processing Systems}, 2024.
\newblock URL \url{https://openreview.net/forum?id=j2wCrWmgMX}.

\bibitem[Qiu \& Miikkulainen(2024)Qiu and Miikkulainen]{semantic_density}
Xin Qiu and Risto Miikkulainen.
\newblock Semantic density: Uncertainty quantification for large language models through confidence measurement in semantic space.
\newblock In \emph{The Thirty-eighth Annual Conference on Neural Information Processing Systems}, 2024.
\newblock URL \url{https://openreview.net/forum?id=LOH6qzI7T6}.

\bibitem[Rafailov et~al.(2023)Rafailov, Sharma, Mitchell, Manning, Ermon, and Finn]{DPO}
Rafael Rafailov, Archit Sharma, Eric Mitchell, Christopher~D Manning, Stefano Ermon, and Chelsea Finn.
\newblock Direct preference optimization: Your language model is secretly a reward model.
\newblock In A.~Oh, T.~Naumann, A.~Globerson, K.~Saenko, M.~Hardt, and S.~Levine (eds.), \emph{Advances in Neural Information Processing Systems}, volume~36, pp.\  53728--53741. Curran Associates, Inc., 2023.
\newblock URL \url{https://proceedings.neurips.cc/paper_files/paper/2023/file/a85b405ed65c6477a4fe8302b5e06ce7-Paper-Conference.pdf}.

\bibitem[Shao et~al.(2024)Shao, Wang, Zhu, Xu, Song, Bi, Zhang, Zhang, Li, Wu, et~al.]{shao2024deepseekmath}
Zhihong Shao, Peiyi Wang, Qihao Zhu, Runxin Xu, Junxiao Song, Xiao Bi, Haowei Zhang, Mingchuan Zhang, YK~Li, Y~Wu, et~al.
\newblock Deepseekmath: Pushing the limits of mathematical reasoning in open language models.
\newblock \emph{arXiv preprint arXiv:2402.03300}, 2024.

\bibitem[Tjandra et~al.(2024)Tjandra, Razzak, Kossen, Handa, and Gal]{SE_Tuning}
Benedict~Aaron Tjandra, Muhammed Razzak, Jannik Kossen, Kunal Handa, and Yarin Gal.
\newblock Fine-tuning large language models to appropriately abstain with semantic entropy.
\newblock In \emph{Neurips Safe Generative AI Workshop 2024}, 2024.
\newblock URL \url{https://openreview.net/forum?id=oYfYgDeSkj}.

\bibitem[Welbl et~al.(2017)Welbl, Liu, and Gardner]{welbl-etal-2017-crowdsourcing}
Johannes Welbl, Nelson~F. Liu, and Matt Gardner.
\newblock Crowdsourcing multiple choice science questions.
\newblock In Leon Derczynski, Wei Xu, Alan Ritter, and Tim Baldwin (eds.), \emph{Proceedings of the 3rd Workshop on Noisy User-generated Text}, pp.\  94--106, Copenhagen, Denmark, September 2017. Association for Computational Linguistics.
\newblock \doi{10.18653/v1/W17-4413}.
\newblock URL \url{https://aclanthology.org/W17-4413/}.

\bibitem[Wen et~al.(2025)Wen, Yao, Feng, Xu, Tsvetkov, Howe, and Wang]{abstention_survey}
Bingbing Wen, Jihan Yao, Shangbin Feng, Chenjun Xu, Yulia Tsvetkov, Bill Howe, and Lucy~Lu Wang.
\newblock Know your limits: A survey of abstention in large language models.
\newblock \emph{Transactions of the Association for Computational Linguistics}, 13:\penalty0 529--556, 2025.
\newblock \doi{10.1162/tacl_a_00754}.
\newblock URL \url{https://aclanthology.org/2025.tacl-1.26/}.

\bibitem[Xia et~al.(2025)Xia, Xu, Zhang, and Liu]{UQ_survey}
Zhiqiu Xia, Jinxuan Xu, Yuqian Zhang, and Hang Liu.
\newblock A survey of uncertainty estimation methods on large language models.
\newblock In Wanxiang Che, Joyce Nabende, Ekaterina Shutova, and Mohammad~Taher Pilehvar (eds.), \emph{Findings of the Association for Computational Linguistics: ACL 2025}, pp.\  21381--21396, Vienna, Austria, July 2025. Association for Computational Linguistics.
\newblock ISBN 979-8-89176-256-5.
\newblock URL \url{https://aclanthology.org/2025.findings-acl.1101/}.

\bibitem[Xu et~al.(2024{\natexlab{a}})Xu, Zhu, Zhang, Ma, Fan, Chen, and Yu]{xurejection}
Hongshen Xu, Zichen Zhu, Situo Zhang, Da~Ma, Shuai Fan, Lu~Chen, and Kai Yu.
\newblock Rejection improves reliability: Training {LLM}s to refuse unknown questions using {RL} from knowledge feedback.
\newblock In \emph{First Conference on Language Modeling}, 2024{\natexlab{a}}.
\newblock URL \url{https://openreview.net/forum?id=lJMioZBoR8}.

\bibitem[Xu et~al.(2024{\natexlab{b}})Xu, Wu, Diao, Liu, Wang, Chen, and Gao]{xu-etal-2024-sayself}
Tianyang Xu, Shujin Wu, Shizhe Diao, Xiaoze Liu, Xingyao Wang, Yangyi Chen, and Jing Gao.
\newblock {S}ay{S}elf: Teaching {LLM}s to express confidence with self-reflective rationales.
\newblock In Yaser Al-Onaizan, Mohit Bansal, and Yun-Nung Chen (eds.), \emph{Proceedings of the 2024 Conference on Empirical Methods in Natural Language Processing}, pp.\  5985--5998, Miami, Florida, USA, November 2024{\natexlab{b}}. Association for Computational Linguistics.
\newblock \doi{10.18653/v1/2024.emnlp-main.343}.
\newblock URL \url{https://aclanthology.org/2024.emnlp-main.343/}.

\bibitem[Xue et~al.(2025)Xue, Mi, Zhu, Wang, Wang, Wang, Yu, Hu, and Wong]{ualign}
Boyang Xue, Fei Mi, Qi~Zhu, Hongru Wang, Rui Wang, Sheng Wang, Erxin Yu, Xuming Hu, and Kam-Fai Wong.
\newblock {UA}lign: Leveraging uncertainty estimations for factuality alignment on large language models.
\newblock In Wanxiang Che, Joyce Nabende, Ekaterina Shutova, and Mohammad~Taher Pilehvar (eds.), \emph{Proceedings of the 63rd Annual Meeting of the Association for Computational Linguistics (Volume 1: Long Papers)}, pp.\  6002--6024, Vienna, Austria, July 2025. Association for Computational Linguistics.
\newblock ISBN 979-8-89176-251-0.
\newblock URL \url{https://aclanthology.org/2025.acl-long.299/}.

\bibitem[Yang et~al.(2024)Yang, Yang, Zhang, Hui, Zheng, Yu, Li, Liu, Huang, Dong, Wei, Lin, Yang, Tu, Zhang, Yang, Yang, Zhou, Lin, Dang, Lu, Bao, Yang, Yu, Li, Xue, Zhang, Zhu, Men, Lin, Li, Xia, Ren, Ren, Fan, Su, Zhang, Wan, Liu, Cui, Zhang, Qiu, Quan, and Wang]{qwen2.5}
An~Yang, Baosong Yang, Beichen Zhang, Binyuan Hui, Bo~Zheng, Bowen Yu, Chengyuan Li, Dayiheng Liu, Fei Huang, Guanting Dong, Haoran Wei, Huan Lin, Jian Yang, Jianhong Tu, Jianwei Zhang, Jianxin Yang, Jiaxin Yang, Jingren Zhou, Junyang Lin, Kai Dang, Keming Lu, Keqin Bao, Kexin Yang, Le~Yu, Mei Li, Mingfeng Xue, Pei Zhang, Qin Zhu, Rui Men, Runji Lin, Tianhao Li, Tingyu Xia, Xingzhang Ren, Xuancheng Ren, Yang Fan, Yang Su, Yi-Chao Zhang, Yunyang Wan, Yuqi Liu, Zeyu Cui, Zhenru Zhang, Zihan Qiu, Shanghaoran Quan, and Zekun Wang.
\newblock Qwen2.5 technical report.
\newblock \emph{arXiv preprint arXiv:2412.15115}, 2024.

\bibitem[Yang et~al.(2025)Yang, Tu, Liu, Wang, Zheng, Zhang, Cui, Chen, He, Wang, et~al.]{yang2025barrel}
Junxiao Yang, Jinzhe Tu, Haoran Liu, Xiaoce Wang, Chujie Zheng, Zhexin Zhang, Shiyao Cui, Caishun Chen, Tiantian He, Hongning Wang, et~al.
\newblock Barrel: Boundary-aware reasoning for factual and reliable lrms.
\newblock \emph{arXiv preprint arXiv:2505.13529}, 2025.

\bibitem[Zhang et~al.(2024)Zhang, Diao, Lin, Fung, Lian, Wang, Chen, Ji, and Zhang]{R-tuning}
Hanning Zhang, Shizhe Diao, Yong Lin, Yi~Fung, Qing Lian, Xingyao Wang, Yangyi Chen, Heng Ji, and Tong Zhang.
\newblock {R}-tuning: Instructing large language models to say {\textquoteleft}{I} don`t know'.
\newblock In Kevin Duh, Helena Gomez, and Steven Bethard (eds.), \emph{Proceedings of the 2024 Conference of the North American Chapter of the Association for Computational Linguistics: Human Language Technologies (Volume 1: Long Papers)}, pp.\  7113--7139, Mexico City, Mexico, June 2024. Association for Computational Linguistics.
\newblock \doi{10.18653/v1/2024.naacl-long.394}.
\newblock URL \url{https://aclanthology.org/2024.naacl-long.394/}.

\bibitem[Zhang et~al.(2023)Zhang, Li, Cui, Cai, Liu, Fu, Huang, Zhao, Zhang, Chen, et~al.]{zhang2023siren}
Yue Zhang, Yafu Li, Leyang Cui, Deng Cai, Lemao Liu, Tingchen Fu, Xinting Huang, Enbo Zhao, Yu~Zhang, Yulong Chen, et~al.
\newblock Siren's song in the ai ocean: a survey on hallucination in large language models.
\newblock \emph{arXiv preprint arXiv:2309.01219}, 2023.

\bibitem[Zhu et~al.(2025)Zhu, Jiang, Wu, Ma, Song, Bai, Lin, Wu, and He]{zhu-etal-2025grait}
Runchuan Zhu, Xinke Jiang, Jiang Wu, Zhipeng Ma, Jiahe Song, Fengshuo Bai, Dahua Lin, Lijun Wu, and Conghui He.
\newblock {GRAIT}: Gradient-driven refusal-aware instruction tuning for effective hallucination mitigation.
\newblock In \emph{Findings of the Association for Computational Linguistics: NAACL 2025}, pp.\  4006--4021, Albuquerque, New Mexico, April 2025. Association for Computational Linguistics.
\newblock ISBN 979-8-89176-195-7.
\newblock \doi{10.18653/v1/2025.findings-naacl.223}.
\newblock URL \url{https://aclanthology.org/2025.findings-naacl.223/}.

\end{thebibliography}
\bibliographystyle{iclr2026_conference}

\newpage
\appendix

\section{Experiment Details}
\label{Hyperparameter_Setup}
\subsection{Dataset}
We test on the following publicly available datasets.
\begin{itemize}[leftmargin=1.2em]
    \item \textbf{Pararel}~\citep{pararel_dataset} is a dataset of factual knowledge and relations that are originally for evaluating masked language models. We follow \citet{R-tuning}, changing it into a question-answering format for generation evaluation. The validation set contains 5,584 QA pairs, which we use for evaluation.
    \item \textbf{TriviaQA}~\citep{joshi-etal-2017-triviaqa} is a large-scale reading comprehension dataset containing over 95,000 trivia-style question–answer pairs. The validation set contains 11,313 QA pairs, which we use for evaluation.
    \item \textbf{Natural Questions (NQ)}~\citep{kwiatkowski-etal-2019-natural} are aggregated queries issued to the Google search engine. All questions of NQ can be answered using the contents of the English Wikipedia. The validation set contains 3,610 QA pairs, which we use for evaluation.
    \item \textbf{SciQ}~\citep{welbl-etal-2017-crowdsourcing} is a crowdsourced dataset of 13,679 science questions. It covers topics from biology, chemistry, physics, and earth science from real educational materials. The validation set contains 1,000 QA pairs, which we use for evaluation.
   
\end{itemize}

\subsection{Details of baselines}
The core idea of existing abstention methods is to fine-tune LLMs to answer the questions within their knowledge scope and abstain from those outside it. It consists of two steps.

First, given a training dataset $D$, these methods partition it into two subsets, the answerable set $D_0$ and the unanswerable set $D_1$. For questions in $D_0$, the model is assumed to be confident about it, and the ground-truth answer is set to be the model’s standard response. For questions in $D_1$, the ground-truth text, it is changed to an abstention expression: ``I don't know''. R-Tuning~\citep{R-tuning} prompts LLMs to answer questions from the training set, then classifies correctly answered questions into $D_0$ and incorrectly answered ones into $D_1$. R-Tuning-U~\citep{R-tuning} and SE-Tuning~\citep{SE_Tuning} use uncertainty to partition the training set. R-Tuning adopts predictive entropy, which clusters lexical equivalent answers together. SE-Tuning adopts semantic entropy, which clusters semantically equivalent answers together. Let $C_1, . . . , C_m$ be the classes that were clustered from sampled responses; the entropy is given by:
\begin{equation}
    Entropy(\mathbf{x}) \approx -\sum_{i=1}^{m} p_{\theta}(C_i \mid \mathbf{x}) \log p_{\theta}(C_i \mid \mathbf{x}) \approx -\sum_{i=1}^{m} \left( \frac{|C_i|}{M} \right) \log \left( \frac{|C_i|}{M} \right),
\label{eq:entropy_approximation}
\end{equation}
where $M$ is the total number of samples. Then, $D_0$ and $D_1$ is obtained by a threshold $\tau$. Then, LLMs are fine-tuned on $D_0$ and $D_1$, learning to minimize the following objective:
\begin{equation}
    \mathcal{L}_{CE}(p_\theta) = -\sum_{x \in {D_0, D_1}} \sum_{t=1}^{|\mathbf{y}^{(x)}|} \log p_\theta(y_t^{(x)} \mid \text{prompt}, \mathbf{x}, \mathbf{y}_{< t}^{(x)}),
    \label{eq:cross_entropy_loss}
\end{equation}

For consistency, GRPO-SE is trained on the same processed dataset as SE-Tuning. Its objective function is Equation~\ref{eq:grpo_obj}, enabling a controlled comparison of reward granularity and abstention behavior.
\subsection{Training Details}
We use 15,000 samples from the Pararel dataset for training. Before training, we remove questions with semantic entropy below 1. These low-entropy samples lack sufficient semantic diversity, which can bias the binary confidence reward and lead the model to abstain too infrequently. Moreover, their simplicity diminishes the relative advantage signal in GRPO training, reducing the effectiveness of policy optimization. In the semantic clustering, we concatenate the question and the answer as the input of the NLI model. we run each experiment with three different random seeds and report the mean performance.

\subsection{Hyperparameters}
For \Ours on Llama3-8B-Instruct and Qwen2.5-7B-Instruct, we use the same hyperparameter setups, which are shown in Table~\ref{table_hyperparameter}.
\begin{table}[h]
\caption{Hyperparameter setups for GRPO trainer.}
\label{table_hyperparameter}
\begin{center} 
\begin{tabular}{ll}
\toprule
\textbf{Parameter} & \textbf{Value} \\
\midrule
\multicolumn{2}{l}{\textit{General Settings}} \\
bf16 & true \\
use\_vllm & true \\
vllm\_device & auto \\
vllm\_enforce\_eager & true \\
vllm\_gpu\_memory\_utilization & 0.7 \\
vllm\_max\_model\_len & 4608 \\
do\_eval & false \\
\midrule
\multicolumn{2}{l}{\textit{Training Configuration}} \\
gradient\_accumulation\_steps & 4 \\
gradient\_checkpointing & true \\
gradient\_checkpointing\_kwargs & use\_reentrant: false \\
learning\_rate & 3.0e-06 \\
lr\_scheduler\_type & cosine\_with\_min\_lr \\
lr\_scheduler\_kwargs & min\_lr\_rate: 0.1 \\
warmup\_ratio & 0.05 \\
num\_train\_epochs & 1 \\
per\_device\_train\_batch\_size & 10 \\
per\_device\_eval\_batch\_size & 10 \\
threshold of abstention  & 5 \\
\midrule
\multicolumn{2}{l}{\textit{Generation Settings}} \\
max\_prompt\_length & 256 \\
max\_completion\_length & 64 \\
num\_generations & 10 \\
temperature & 1.0 \\
seed & 42, 32, 22 \\
\midrule
\multicolumn{2}{l}{\textit{Logging and Saving}} \\
log\_completions & true \\
log\_level & info \\
logging\_first\_step & true \\
logging\_steps & 1 \\
logging\_strategy & steps \\
save\_strategy & steps \\
save\_steps & 50 \\
report\_to & wandb \\
\midrule
\multicolumn{2}{l}{\textit{Reward Configuration}} \\
reward\_funcs & confidence, format, accuracy (semantic) \\
reward\_weights & 1.0 2.0, 4.0 \\
\bottomrule
\end{tabular}
\end{center}
\label{tab:hypers}
\end{table}

\section{Reliability Metric}
\label{appendix_F1_rel}

\subsection{Analysis for Reliability Score}\label{app:reliability_score}

\citet{xurejection} propose the Reliability Score (RS), which encourages LLMs to offer maximal assistance while minimizing hallucinated errors. RS is defined as a weighted sum of accuracy and truthful rate, which is used by current studies~\citep{xurejection,when_to_speak,yang2025barrel}: $\text{RS}(\alpha) = \alpha \times \text{Truth.} + (1 - \alpha) \times \text{Acc.}$, the weight $\alpha$ is answering rate. 

The components are:
\begin{align*}
    N &= N_1+N_2+N_3+N_4+N_5 \\
    \text{Acc.} &= \frac{N_1}{N} \\
    \text{Truth.} &= \frac{N_1+N_3+N_5}{N} \\
    \alpha &= 1 - \frac{N_3+N_5}{N} = \frac{N_1+N_2+N_4}{N}
\end{align*}
Substituting these into the RS formula:
\begin{align*}
    RS &= \left(\frac{N_1+N_2+N_4}{N}\right) \left(\frac{N_1+N_3+N_5}{N}\right) + \left(1 - \frac{N_1+N_2+N_4}{N}\right) \left(\frac{N_1}{N}\right) \\
    &= \left(\frac{N_1+N_2+N_4}{N}\right) \left(\frac{N_1+N_3+N_5}{N}\right) + \left(\frac{N_3+N_5}{N}\right) \left(\frac{N_1}{N}\right) \\
    &= \frac{(N_1+N_2+N_4)(N_1+N_3+N_5) + N_1(N_3+N_5)}{N^2}
\end{align*}

The partial derivative of RS with respect to $N_4$ is:
\begin{equation}
    \frac{\partial \text{RS}}{\partial N_4} = \frac{N_3 + N_5 - (N_1 + N_2 + N_4)}{N^2}
\end{equation}
The sign of this derivative is variable and contingent on the specific distribution of counts among the categories. It is not guaranteed to be negative. For instance, consider a perfect model evaluated on a dataset composed solely of unanswerable questions ($N_1=N_2=N_3=0$, with an initial state of $N_4=0$ and $N_5=N$). The derivative at this point is:
\begin{equation}
    \frac{\partial \text{RS}}{\partial N_4} \bigg|_{N_4=0} = \frac{N_5 - N_4}{N^2} = \frac{N - 0}{N^2} = \frac{1}{N} > 0
\end{equation}
A positive derivative demonstrates that RS \textbf{perversely incentivizes} a perfect model to start making mistakes. An initial increase in $N_4$ from zero leads to an \textit{increase} in the RS score. This behavior is antithetical to the fundamental goal of a reliability metric.

\subsection{Analysis for the proposed reliability F1}

The proposed reliability $\text{F1}_{rel}$ is defined as:
$$ \text{F1}_{rel} = \frac{2}{\frac{1}{\text{F1}_{ans}} + \frac{1}{\text{F1}_{abs}}} $$

First, we calculate the sum of the reciprocals:
\begin{align*}
    \frac{1}{\text{F1}_{ans}} + \frac{1}{\text{F1}_{abs}} &= \frac{2N_1+2N_2+N_3+N_4}{2N_1} + \frac{N_3+N_4+2N_5}{2N_5} \\
    &= \frac{N_5(2N_1+2N_2+N_3+N_4) + N_1(N_3+N_4+2N_5)}{2N_1N_5} \\
    &= \frac{2N_1N_5+2N_2N_5+N_3N_5+N_4N_5 + N_1N_3+N_1N_4+2N_1N_5}{2N_1N_5} \\
    &= \frac{4N_1N_5+2N_2N_5+N_1N_3+N_1N_4+N_3N_5+N_4N_5}{2N_1N_5}
\end{align*}

Finally, substitute this result back into the formula for $\text{F1}_{rel.}$:
\begin{align*}
    \text{F1}_{rel} &= \frac{2}{\frac{4N_1N_5+2N_2N_5+N_1N_3+N_1N_4+N_3N_5+N_4N_5}{2N_1N_5}} \\
    &= \frac{4N_1N_5}{4N_1N_5+2N_2N_5+N_1N_3+N_1N_4+N_3N_5+N_4N_5}
\end{align*}

In stark contrast, the partial derivative of $\text{F1}_{rel}$ with respect to $N_4$ is:
\begin{equation}
    \frac{\partial {\text{F1}_{rel}}}{\partial N_4} = -\frac{\text{F1}_{rel}^2}{4} \left( \frac{1}{N_1} + \frac{N_3 + N_4 + N_5}{N_5^2} \right)
\end{equation}
The sign of this derivative is \textbf{consistently non-positive}. We can analyze its components to confirm this:
\begin{itemize}
    \item The term $\text{F1}_{rel}^2$ is always non-negative.
    \item The term within the parentheses, $\left( \frac{1}{N_1} + \frac{N_3 + N_4 + N_5}{N_5^2} \right)$, is a sum of non-negative quantities and is strictly positive for any non-trivial case where the metric is defined ($N_1 > 0$ and $N_5 > 0$).
    \item The leading factor of $-\frac{1}{4}$ ensures the entire expression is non-positive.
\end{itemize}
This non-positive derivative signifies the ideal behavior. It mathematically guarantees that any increase in the critical error count $N_4$ will \textbf{always lead to a decrease or no change} in the $\text{F1}_{rel}$ score. The metric consistently and correctly penalizes unreliability.

\paragraph{Conclusion.} The derivative analysis provides a rigorous mathematical basis for concluding that $\text{F1}_{rel}$ is a superior metric for evaluating model reliability. While the RS metric can fail catastrophically by rewarding incorrect behavior, the $\text{F1}_{rel}$ metric provides a robust, reliable, and interpretable gradient that always punishes critical errors. This makes $\text{F1}_{rel}$ a far more suitable and trustworthy measure for developing reliable AI systems.

\subsection{Special Case}
To highlight the superiority of our proposed metric $\text{F1}_{rel}$ (the harmonic mean of $\text{F1}_{ans}$ and $\text{F1}_{abs}$) over the Reliability Score (RS), we analyze a specific, illustrative scenario: a dataset composed exclusively of unanswerable questions. A robust reliability metric must correctly identify and reward a model that abstains from such questions, while penalizing a model that makes incorrect guesses. Consider a dataset that contains only unanswerable questions. This implies:
$$ N_1 = 0, \quad N_2 = 0, \quad N_3 = 0 $$
the total number of samples is therefore $N = N_4 + N_5$.

We evaluate two distinct models on this dataset:
\begin{itemize}
    \item \textbf{Model A (The Perfect Abstainer):} This model performs perfectly by correctly identifying all questions as unanswerable and abstaining every time, defined by $N_5 = N, N_4 = 0$.
    \item \textbf{Model B (The Reckless Guesser):} This model is flawed. It abstains half the time but provides an incorrect answer for the other half, defined by $N_5 = 0.5N,N_4 = 0.5N $.
\end{itemize}
\textbf{Our expectation:} A valid metric should award Model A a perfect (or maximum) score and Model B a significantly lower score.

\textbf{Analysis of the Reliability Score (RS)}
The simplified formula for RS in this case is:
\begin{align*}
    \text{RS} = (\alpha \cdot \text{Cov}) + ((1-\alpha) \cdot \text{Acc}) = (\frac{N_4}{N} \cdot \frac{N_5}{N}) + ((1-\frac{N_4}{N}) \cdot 0) = \frac{N_4 N_5}{N^2}.
\end{align*}

For Model A:
\begin{align*}
    \text{RS}_A &= \frac{(0)(N)}{N^2} = 0
\end{align*}
For Model B:
\begin{align*}
    \text{RS}_B &= \frac{(0.5N)(0.5N)}{N^2} = \frac{0.25N^2}{N^2} = 0.25
\end{align*}
RS erroneously assigns Model A a score of 0 and Model B a higher score of 0.25.

\textbf{Evaluation of the proposed $\text{F1}_{rel}$ Metric}

In this scenario, $\text{F1}_{rel}$ depends solely on $\text{F1}_{abs}$, reflecting abstention performance:
\begin{align*}
    \text{F1}_{rel} = \text{F1}_{abs} &= \frac{2N_5}{N_3+N_4+2N_5}
\end{align*}
For Model A:
\begin{align*}
    \text{F1}_{absA} &= \frac{2(N)}{0+0+2(N)} = \frac{2N}{2N} = \mathbf{1}
\end{align*}
This assigns a perfect score of 1, reflecting optimal performance. For Model B:
\begin{align*}
    \text{F1}_{absB} &= \frac{2(0.5N)}{0+0.5N+2(0.5N)} = \frac{N}{0.5N+N} = \frac{N}{1.5N} = \frac{2}{3} \approx 0.67
\end{align*}
This appropriately penalizes Model B for incorrect answers.

\textbf{Conclusion}

The Reliability Score (RS) exhibits a critical flaw by rewarding incorrect answers on unanswerable questions, contradicting the principles of reliability. In contrast, $\text{F1}_{rel}$, via its $\text{F1}_{abs}$ component, accurately assigns Model A a perfect score and penalizes Model B, demonstrating its robustness and suitability as a reliability metric.

\section{Evaluation Prompt}
\label{Evaluation_Prompt}

We use a 6-shot prompt across all methods. For the ICL baseline, the prompt consists of six examples of direct answering. In contrast, prompts for abstention-aware methods include a balanced mix of three answering examples and three abstention examples.
\definecolor{almond}{RGB}{239, 222, 205}

\begin{quote}
    \begin{tcolorbox}[ title=\textbf{ICL Prompt}]
    \label{temp:icl}
        
        Answer the following questions as briefly as possible.\\
        \\
        Question: {\texttt{\{demo question 1\}}}\\
        Answer: {\texttt{\{demo answer 1\}}}\\
        \\
        Question: {\texttt{\{demo question 2\}}}\\
        Answer: {\texttt{\{demo answer 2\}}}\\
        \\
        ...\\
        \\
        Question: {\texttt{\{input question\}}}\\
        Answer: \\
    \end{tcolorbox}
    
    \begin{tcolorbox}[title=\textbf{ICL-IDK Prompt}]
    \label{temp:icl-idk}
        Answer the following questions as briefly as possible. If you don't know the answer, please simply say ``I don't know.''\\
        \\
        Question: {\texttt{\{demo question 1\}}}\\
        Answer: {\texttt{\{demo answer 1\}}}\\
        \\
        Question: {\texttt{\{demo question 2\}}}\\
        Answer: {\texttt{I don't know.}}\\
        \\
        ...\\
        \\
        Question: {\texttt{\{input question\}}}\\
        Answer: \\
    \end{tcolorbox}
    
    \begin{tcolorbox}[ title=\textbf{ICL-Unsure Prompt}]
    \label{temp:icl-cot}
        You are a helpful and truthful AI Assistant. You should answer the question as briefly as possible, then present your confidence. If you are sure about your answer, please say ``I am sure'' after your answer; otherwise, say ``I am unsure''.\\
        \\
        Question: {\texttt{\{demo question 1\}}}\\
        Answer: {\texttt{\{demo answer 1\}}} I am sure.\\
        \\
        Question: {\texttt{\{demo question 2\}}}\\
        Answer: {\texttt{\{demo answer 2\}}} I am unsure.\\
        \\
        ...\\
        \\
        Question: {\texttt{\{input\_question\}}}\\
        Answer: \\
    \end{tcolorbox}

    \begin{tcolorbox}[title=\textbf{R-Tuning, R-Tuning-U and SE-Tuning Prompt}]
    \label{temp:r-tuning}
        Answer the following questions as briefly as possible. If you don't know the answer, please simply say ``I don't know.''\\
        \\
        Question: {\texttt{\{demo question 1\}}}\\
        Answer: {\texttt{\{demo answer 1\}}}\\
        \\
        Question: {\texttt{\{demo question 2\}}}\\
        Answer: {\texttt{I don't know.}}\\
        \\
        ...\\
        \\
        Question: {\texttt{\{input question\}}}\\
        Answer: \\
    \end{tcolorbox}

    \begin{tcolorbox}[ title=\textbf{GRPO-SE and \Ours Prompt}]
    \label{temp:icl-cot2}
        You are a helpful and truthful AI Assistant that provides reponses include answer and confidence. You first answer the question as briefly as possible enclosed by \verb|<answer>| and \verb|</answer>| and then provide your confidence in sure or unsure about the answer enclosed by \verb|<confidence>| and \verb|</confidence>|. Respond in the following format: \\ \verb|<answer>...</answer>| \\ 
        \verb|<confidence> sure or unsure </confidence>|\\
        \\
        Question: \verb|{demo question 1}|\\
        Answer: \verb|<answer> {demo answer 1} </answer>| \\
                \verb|<confidence> sure </confidence>|
        \\ \\
        Question: \verb|{demo question 2}|\\
        Answer: \verb|<answer> {demo answer 2} </answer>| \\
                \verb|<confidence> unsure </confidence>|
        \\ \\
        ...\\
        \\
        Question: {\texttt{\{input question\}}}\\
        Answer: \\
    \end{tcolorbox}
\end{quote}

\section{Case Study}
\label{appendix_case_study}
We present detailed examples using Llama3-8B-Instruct in Table~\ref{table_case_study}. In question 1, both ICL-IDK and GRPO-SE exhibit over-abstention. While ICL achieves an accuracy of 70\%, ICL-IDK generates two correct answers, one hallucinated response, and seven abstentions. GRPO-SE produces six correct answers but discards five of them. In contrast, \Ours successfully retains all correct answers while discarding all incorrect ones. In question 2, GRPO-SE demonstrates overaggressive behavior by retaining all incorrect answers, whereas \Ours appropriately discards three incorrect responses, achieving calibrated post-hoc abstention.
\begin{table}
\caption{Case studies of different methods on Llama3-8B-Instruct. We sample 10 answers for each question with the temperature of 0.2.}
\label{table_case_study}
\begin{center}
\small
\begin{tabular}{|l|l|l|l|} 
\toprule
\multicolumn{4}{|l|}{\begin{tabular}[c]{@{}l@{}}Question1: The world's largest marketer of fruit juices, what is the juice arm of the Coca Cola company? \\Answer: maid\end{tabular}}       \\ 
\midrule
\multicolumn{1}{|c|}{\textbf{ICL}}              & \multicolumn{1}{c|}{\textbf{ICL-IDK}} & \multicolumn{1}{c|}{\textbf{GRPO-SE}}                                                                                                                                                            & \multicolumn{1}{c|}{\textbf{\textbf{\Ours}}}                                                  \\ 
\midrule
minute maid                                     & maid                                  & \begin{tabular}[c]{@{}l@{}}\textless{}answer\textgreater{} maid \textless{}answer\textgreater{} \\\textless{}confidence\textgreater{} sure \textless{}/confidence\textgreater{}\end{tabular}     & \begin{tabular}[c]{@{}l@{}}\textless{}answer\textgreater{} maid \textless{}answer\textgreater{} \\\textless{}confidence\textgreater{} sure \textless{}/confidence\textgreater{}\end{tabular}         \\ 
\hline
minute maid                                     & maid                                  & \begin{tabular}[c]{@{}l@{}}\textless{}answer\textgreater{} maid \textless{}answer\textgreater{} \\\textless{}confidence\textgreater{} unsure \textless{}/confidence\textgreater{}\end{tabular}   & \begin{tabular}[c]{@{}l@{}}\textless{}answer\textgreater{} maid \textless{}answer\textgreater{} \\\textless{}confidence\textgreater{} sure \textless{}/confidence\textgreater{}\end{tabular}         \\ 
\hline
minute maid                                     & simply                                & \begin{tabular}[c]{@{}l@{}}\textless{}answer\textgreater{} maid \textless{}answer\textgreater{} \\\textless{}confidence\textgreater{} unsure \textless{}/confidence\textgreater{}\end{tabular}   & \begin{tabular}[c]{@{}l@{}}\textless{}answer\textgreater{} maid \textless{}answer\textgreater{} \\\textless{}confidence\textgreater{} sure \textless{}/confidence\textgreater{}\end{tabular}         \\ 
\hline
minute maid                                     & I don't know.                          & \begin{tabular}[c]{@{}l@{}}\textless{}answer\textgreater{} maid \textless{}answer\textgreater{} \\\textless{}confidence\textgreater{} unsure \textless{}/confidence\textgreater{}\end{tabular}   & \begin{tabular}[c]{@{}l@{}}\textless{}answer\textgreater{} maid \textless{}answer\textgreater{} \\\textless{}confidence\textgreater{} sure \textless{}/confidence\textgreater{}\end{tabular}         \\ 
\hline
minute maid                                     & I don't know.                         & \begin{tabular}[c]{@{}l@{}}\textless{}answer\textgreater{} maid \textless{}answer\textgreater{} \\\textless{}confidence\textgreater{} unsure \textless{}/confidence\textgreater{}\end{tabular}   & \begin{tabular}[c]{@{}l@{}}\textless{}answer\textgreater{} maid \textless{}answer\textgreater{} \\\textless{}confidence\textgreater{} sure \textless{}/confidence\textgreater{}\end{tabular}         \\ 
\hline
minute maid                                     & I don't know.                         & \begin{tabular}[c]{@{}l@{}}\textless{}answer\textgreater{} maid \textless{}answer\textgreater{} \\\textless{}confidence\textgreater{} unsure \textless{}/confidence\textgreater{}\end{tabular}   & \begin{tabular}[c]{@{}l@{}}\textless{}answer\textgreater{} maid \textless{}answer\textgreater{} \\\textless{}confidence\textgreater{} sure \textless{}/confidence\textgreater{}\end{tabular}         \\ 
\hline
minute maid                                     & I don't know.                         & \begin{tabular}[c]{@{}l@{}}\textless{}answer\textgreater{} simply \textless{}answer\textgreater{} \\\textless{}confidence\textgreater{} unsure \textless{}/confidence\textgreater{}\end{tabular} & \begin{tabular}[c]{@{}l@{}}\textless{}answer\textgreater{} simply \textless{}answer\textgreater{} \\\textless{}confidence\textgreater{} unsure \textless{}/confidence\textgreater{}\end{tabular}     \\ 
\hline
simply                                          & I don't know.                         & \begin{tabular}[c]{@{}l@{}}\textless{}answer\textgreater{} simply \textless{}answer\textgreater{} \\\textless{}confidence\textgreater{} unsure \textless{}/confidence\textgreater{}\end{tabular} & \begin{tabular}[c]{@{}l@{}}\textless{}answer\textgreater{} simply \textless{}answer\textgreater{} \\\textless{}confidence\textgreater{} unsure \textless{}/confidence\textgreater{}\end{tabular}     \\ 
\hline
fanta                                           & I don't know.                         & \begin{tabular}[c]{@{}l@{}}\textless{}answer\textgreater{} fanta \textless{}answer\textgreater{} \\\textless{}confidence\textgreater{} unsure \textless{}/confidence\textgreater{}\end{tabular}  & \begin{tabular}[c]{@{}l@{}}\textless{}answer\textgreater{} fanta \textless{}answer\textgreater{} \\\textless{}confidence\textgreater{} unsure \textless{}/confidence\textgreater{}\end{tabular}      \\ 
\hline
\begin{tabular}[c]{@{}l@{}}fanta\\\end{tabular} & I don't know.                         & \begin{tabular}[c]{@{}l@{}}\textless{}answer\textgreater{} fanta \textless{}answer\textgreater{} \\\textless{}confidence\textgreater{} unsure \textless{}/confidence\textgreater{}\end{tabular}  & \begin{tabular}[c]{@{}l@{}}\textless{}answer\textgreater{} fanta \textless{}answer\textgreater{} \\\textless{}confidence\textgreater{} unsure \textless{}/confidence\textgreater{}\end{tabular}      \\ 
\midrule
\multicolumn{4}{|l|}{\begin{tabular}[c]{@{}l@{}}Question2: Which mythical beings were said to inhabit the slopes of Mount Etna? \\Answer: Cyclops\end{tabular}}                                                                                                                                                                                                                                                                                                                                    \\ 
\midrule
\multicolumn{1}{|c|}{\textbf{ICL}}              & \multicolumn{1}{c|}{\textbf{ICL-IDK}} & \multicolumn{1}{c|}{\textbf{GRPO-SE}}                                                                                                                                                            & \multicolumn{1}{c|}{\textbf{\textbf{\Ours}}}                                                                                                                                                                   \\ 
\midrule
Cyclops                                         & Cyclops                               & \begin{tabular}[c]{@{}l@{}}\textless{}answer\textgreater{} Cyclops \textless{}/answer\textgreater{}\\\textless{}confidence\textgreater{} sure \textless{}/confidence\textgreater{}\end{tabular}  & \begin{tabular}[c]{@{}l@{}}\textless{}answer\textgreater{} Cyclops \textless{}/answer\textgreater{}\\\textless{}confidence\textgreater{} sure \textless{}/confidence\textgreater{}\end{tabular}      \\ 
\hline
Cyclops                                         & Cyclops                               & \begin{tabular}[c]{@{}l@{}}\textless{}answer\textgreater{} Cyclops \textless{}/answer\textgreater{} \\\textless{}confidence\textgreater{} sure \textless{}/confidence\textgreater{}\end{tabular} & \begin{tabular}[c]{@{}l@{}}\textless{}answer\textgreater{} Cyclops \textless{}/answer\textgreater{} \\\textless{}confidence\textgreater{} sure \textless{}/confidence\textgreater{}\end{tabular}     \\ 
\hline
Cyclops                                         & Cyclops                               & \begin{tabular}[c]{@{}l@{}}\textless{}answer\textgreater{} Cyclops \textless{}/answer\textgreater{} \\\textless{}confidence\textgreater{} sure \textless{}/confidence\textgreater{}\end{tabular} & \begin{tabular}[c]{@{}l@{}}\textless{}answer\textgreater{} Cyclops \textless{}/answer\textgreater{} \\\textless{}confidence\textgreater{} sure \textless{}/confidence\textgreater{}\end{tabular}     \\ 
\hline
Cyclops                                         & Cyclops                               & \begin{tabular}[c]{@{}l@{}}\textless{}answer\textgreater{} Cyclops \textless{}/answer\textgreater{}\\\textless{}confidence\textgreater{} sure \textless{}/confidence\textgreater{}\end{tabular}  & \begin{tabular}[c]{@{}l@{}}\textless{}answer\textgreater{} Cyclops \textless{}/answer\textgreater{}\\\textless{}confidence\textgreater{} sure \textless{}/confidence\textgreater{}\end{tabular}      \\ 
\hline
Cyclops                                         & Cyclops                               & \begin{tabular}[c]{@{}l@{}}\textless{}answer\textgreater{} Cyclops \textless{}/answer\textgreater{} \\\textless{}confidence\textgreater{} sure \textless{}/confidence\textgreater{}\end{tabular} & \begin{tabular}[c]{@{}l@{}}\textless{}answer\textgreater{} Cyclops \textless{}/answer\textgreater{} \\\textless{}confidence\textgreater{} sure \textless{}/confidence\textgreater{}\end{tabular}     \\ 
\hline
Cyclops                                         & Cyclops                               & \begin{tabular}[c]{@{}l@{}}\textless{}answer\textgreater{} Cyclops \textless{}/answer\textgreater{} \\\textless{}confidence\textgreater{} sure \textless{}/confidence\textgreater{}\end{tabular} & \begin{tabular}[c]{@{}l@{}}\textless{}answer\textgreater{} Cyclops \textless{}/answer\textgreater{} \\\textless{}confidence\textgreater{} sure \textless{}/confidence\textgreater{}\end{tabular}     \\ 
\hline
Cyclops                                         & I don't know.                         & \begin{tabular}[c]{@{}l@{}}\textless{}answer\textgreater{} Cyclops \textless{}/answer\textgreater{} \\\textless{}confidence\textgreater{} sure \textless{}/confidence\textgreater{}\end{tabular} & \begin{tabular}[c]{@{}l@{}}\textless{}answer\textgreater{} Cyclops \textless{}/answer\textgreater{} \\\textless{}confidence\textgreater{} sure \textless{}/confidence\textgreater{}\end{tabular}     \\ 
\hline
Centaurs                                        & I don't know.                         & \begin{tabular}[c]{@{}l@{}}\textless{}answer\textgreater{} Centaurs \textless{}/answer\textgreater{}\\\textless{}confidence\textgreater{} sure \textless{}/confidence\textgreater{}\end{tabular}  & \begin{tabular}[c]{@{}l@{}}\textless{}answer\textgreater{} Centaurs \textless{}/answer\textgreater{}\\\textless{}confidence\textgreater{} unsure \textless{}/confidence\textgreater{}\end{tabular}   \\ 
\hline
Centaurs   & I don't know.                        & \begin{tabular}[c]{@{}l@{}}\textless{}answer\textgreater{} Centaurs \textless{}/answer\textgreater{} \\\textless{}confidence\textgreater{} sure \textless{}/confidence\textgreater{}\end{tabular} & \begin{tabular}[c]{@{}l@{}}\textless{}answer\textgreater{} Centaurs \textless{}/answer\textgreater{} \\\textless{}confidence\textgreater{} unsure \textless{}/confidence\textgreater{}\end{tabular}  \\ 
\hline
Centaurs                                        & Centaurs                              & \begin{tabular}[c]{@{}l@{}}\textless{}answer\textgreater{} Centaurs \textless{}/answer\textgreater{} \\\textless{}confidence\textgreater{} sure \textless{}/confidence\textgreater{}\end{tabular} & \begin{tabular}[c]{@{}l@{}}\textless{}answer\textgreater{} Centaurs \textless{}/answer\textgreater{} \\\textless{}confidence\textgreater{} unsure \textless{}/confidence\textgreater{}\end{tabular}  \\
\bottomrule
\end{tabular}
\end{center}

\end{table}

\section{Computation Cost}
R-Tuning~\citep{R-tuning} trains LLMs to respond with ``I don't know'' explicitly when feeded a uncertain question. In contrast, \Ours enables self-reflective abstention or answering, at the cost of increased computation. We compare ICL, R-Tuning, and \Ours in terms of the average generation length of all questions in the test sets and the generation time on 100 randomly sampled questions using an NVIDIA L40 GPU. As shown in Table~\ref{table_computation_cost}, R-Tuning exhibits slightly slower generation speed than ICL, though in certain cases it performs comparably or even faster. In contrast, \Ours incurs a 3-4x increase in generation time relative to ICL. This overhead arises from the structured output format: for answered questions, our method introduces additional tokens for tags and confidence expression; for abstained questions, it further generates a full-length answer to justify the refusal, thereby increasing computational cost.

\begin{table}[htbp]
\centering
\caption{The results of the computation cost. Left: average number of tokens of the model's response. Right: average generation time of 100 random questions.}
\label{table_computation_cost}
\begin{center}
\begin{tabularx}{\textwidth}{l|XXXX|XXXX}
\toprule
\multirow{2}{*}{\begin{tabular}[c]{@{}l@{}}\textbf{Model}\\\end{tabular}} & \multicolumn{4}{c|}{\textbf{Avg. tokens }}                         & \multicolumn{4}{c}{\textbf{Avg. inference time (s) }}               \\ 
\cmidrule{2-9}
                                                                          & \textbf{Pararel} & \textbf{TriviaQA} & \textbf{NQ} & \textbf{SciQ} & \textbf{Pararel} & \textbf{TriviaQA} & \textbf{NQ} & \textbf{SciQ}  \\ 
\midrule
\multicolumn{9}{c}{\textbf{Llama3-8b-Instruct }}                                                                                                                                                                     \\ 
\midrule
\textbf{ICL}                                                              & 2.86             & 3.93              & 5.14        & 3.47          & 7.75             & 11.93             & 12.36       & 10.39          \\
\textbf{R-Tuning}                                                          & 5.33             & 5.74              & 6.47        & 3.97          & 13.84            & 16.14             & 15.31       & 20.77          \\
\textbf{\Ours}                                                             & 15.83            & 16.82             & 17.70       & 16.06         & 31.79            & 35.29             & 35.13       & 32.82          \\ 
\midrule
\multicolumn{9}{c}{\textbf{Qwen2.5-7B-Instruct }}                                                                                                                                                                    \\ 
\midrule
\textbf{ICL}                                                              & 3.36             & 5.54              & 12.44       & 3.64          & 12.23            & 12.49             & 19.49       & 11.04          \\
\textbf{R-Tuning}                                                          & 4.35             & 4.92              & 5.84        & 3.41          & 13.44            & 15.54             & 16.62       & 20.73          \\
\textbf{\Ours}                                                             & 12.17            & 16.84             & 19.36       & 15.29         & 42.69            & 44.54             & 54.20       & 42.49          \\
\bottomrule
\end{tabularx}
\end{center}
\end{table}

\section{Limitations}
Our method has some limitations. First, we must manually set the threshold $\tau$ during GRPO training. Second, the $\text{F1}_{ans}$, $\text{F1}_{abs}$, and $\text{F1}_{rel}$ metrics are only suitable for evaluating alignment methods, which means that we must know the knowledge that the model possesses before alignment. Future evaluation metrics for LLMs should encourage models to acknowledge their knowledge gaps rather than guessing answers. Our method teaches LLMs to first generate an answer and then express the abstention decision, which requires more computation than directly abstaining from answering.

\end{document}